\pdfoutput=1

\documentclass[11pt]{article}

\usepackage[final]{acl}

\usepackage{times}
\usepackage{latexsym}

\usepackage[T1]{fontenc}

\usepackage[utf8]{inputenc}

\usepackage{microtype}

\usepackage{inconsolata}

\usepackage{graphicx}

\usepackage{tikz}
\usepackage[most]{tcolorbox} 
\usepackage{xcolor}
\usepackage[export]{adjustbox}
\usepackage{multirow}
\usepackage{multicol}
\usepackage{enumitem}
\usepackage{textcomp}

\tcbset{
    boxsep=1pt
}

\newtcolorbox{smallsepTcolorbox}[1]{
    arc=5pt,
    boxsep=1mm,
    breakable
}

\newtcolorbox{breakableTcolorbox}[1]{
    arc=5pt,
    fonttitle=\sffamily\bfseries\small, 
    title=#1, 
    breakable
}

\newtcolorbox{breakableTcolorboxSampleQuestion}[1]{
    arc=5pt,
    fonttitle=\sffamily\bfseries\small,
    colback=blue!5!white,colframe=blue!50!black,
    title=#1,
    breakable
}
\newtcolorbox{breakableTcolorboxGTAnswer}[1]{
    arc=5pt,
    fonttitle=\sffamily\bfseries\small,
    colback=green!5!white,colframe=green!50!black,
    title=#1,
    breakable
}
\newtcolorbox{breakableTcolorboxInitPredict}[1]{
    arc=5pt,
    fonttitle=\sffamily\bfseries\small,
    colback=purple!5!white,colframe=purple!50!black,
    title=#1,
    breakable
}
\newtcolorbox{breakableTcolorboxOptPredict}[1]{
    arc=5pt,
    fonttitle=\sffamily\bfseries\small,
    colback=magenta!5!white,colframe=magenta!50!black,
    title=#1,
    breakable
}

\title{Optimization before Evaluation: Evaluation with Unoptimized Prompts Can be Misleading}

\author{Nicholas Sadjoli\textsuperscript{1}, Tim Siefken\textsuperscript{1}, Atin Ghosh\textsuperscript{1}, Yifan Mai\textsuperscript{2}, Daniel Dahlmeier\textsuperscript{1} \\
        \textsuperscript{1}SAP \ \ \textsuperscript{2}Stanford University\\
        \small \texttt{\{nicholas.sadjoli,tim.siefken,atin.ghosh,d.dahlmeier\}@sap.com} \ \ \ \ \texttt{yifan@cs.stanford.edu}} 

\begin{document}
\maketitle

\begin{abstract}
Current Large Language Model (LLM) evaluation frameworks utilize the same static prompt template across all models under evaluation. This differs from the common industry practice of using prompt optimization (PO) techniques to optimize the prompt for each model to maximize application performance. In this paper, we investigate the effect of PO towards LLM evaluations. Our results on public academic and internal industry benchmarks show that PO greatly affects the final ranking of models. This highlights the importance of practitioners performing PO per model when conducting evaluations to choose the best model for a given task.

\end{abstract}

\section{Introduction} \label{sec_1:introduction}
Due to recent advances in their capabilities and performance, Large Language Models (LLMs) are now being integrated into many real-world applications. However, selecting the optimal LLM for an application is a complicated task that requires evaluating multiple models on a variety of metrics, such as accuracy, consistency, and reliability. Benchmarking frameworks have been developed to address this issue and to systematically find the best model \cite{saini-etal-2025-llm_evaluate-industry_focused_evaluation_tool_for_llms,liang2023-helm-paper,2024-eval-harness}. However, these benchmarks share the common limitation of using a static prompt template when testing across different models \cite{liang2023-helm-paper, srivastava2023-bigbench, dalvi-etal-2024-llmebench}. 

This makes most benchmarks almost entirely \emph{model-centric}: the model is treated as the \emph{interface} and evaluation results only depend on the models' capabilities of `understanding' and completing the task based on the same prompt instruction. However, from an \emph{application-centric} perspective, this approach has some drawbacks. It is well known that prompt quality and style affect a model's instruction following capability and overall performance \cite{pryzant2023-automatic_prompt_optimization_gradient, zhou2023instructionfollowingevaluationlargelanguage, acl2024-instruction-following-behavior-shift-wu-etal, cheng2024-blackbox_prompt_optimization-align_llm_wo_modeltraining, wan2024-instructions_vs_exemplars-auto_prompt_optimization}. This means that the prompts are also variables that can be optimized to achieve maximum application performance and should be considered as part of the model testing. 

The recent development of prompt optimization (PO) methodologies has given us methods for automatically improving the prompt for a given model and task, based on a small number of training samples \cite{pryzant2023-automatic_prompt_optimization_gradient, cheng2024-blackbox_prompt_optimization-align_llm_wo_modeltraining} - which can also include optimized exemplars \cite{wan2024-instructions_vs_exemplars-auto_prompt_optimization}. This can greatly improve the performance and instruction-following capabilities of a model \cite{lu2024-fipo-freeform_instruction_oriented_prompt}. Thus, it seems logical to include PO for application-centric LLM evaluations. However, to the best of our knowledge, PO has not been adopted in any existing benchmarking framework.

In this paper, we investigate the effect of PO in application-centric LLM evaluation. Our experiments on academic and industry benchmarks reveal the following key observations:
\begin{enumerate}
    \item PO generally improves the instruction-following capabilities and performance of models. While performance may decrease for some models in specific use cases, PO generally results in a higher overall performance for a given task. 
    \item PO can change the relative performance rankings of models and should therefore be used for application-centric evaluations when the goal is to pick the best model for a given task. 
    \item Models have different levels of sensitivity to PO, depending on the tasks and data. 
 \end{enumerate}

 \begin{figure*}[htbp]
    \hspace{11mm}
    \includegraphics[width=0.195\linewidth]{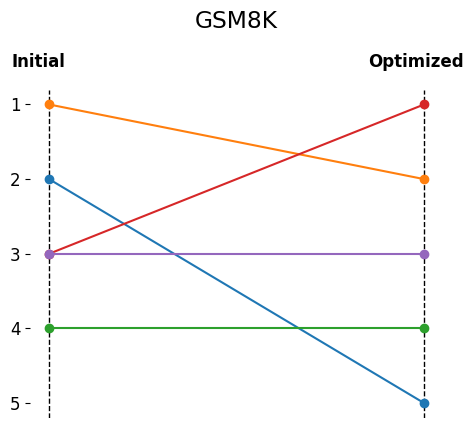} \hspace{1mm}
    \includegraphics[width=0.195\linewidth]{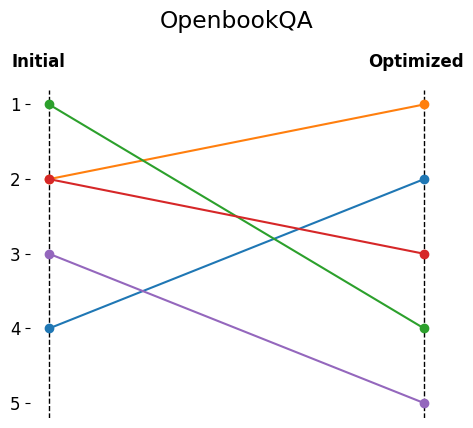} \hspace{1mm}
    \raisebox{0.03\height}{\includegraphics[width=0.265\linewidth]{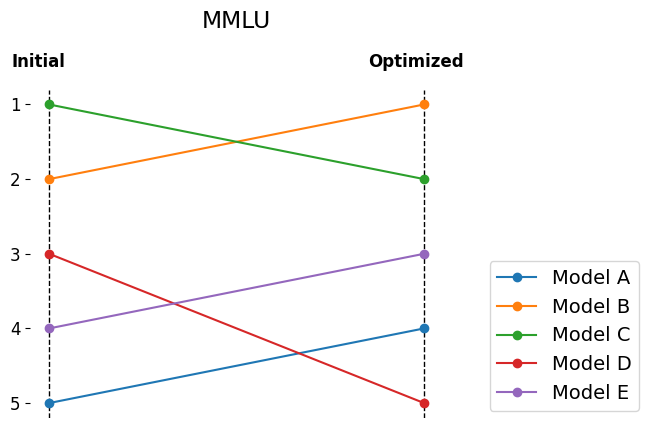}} 
    \hspace{11mm}
    \includegraphics[width=0.195\linewidth]{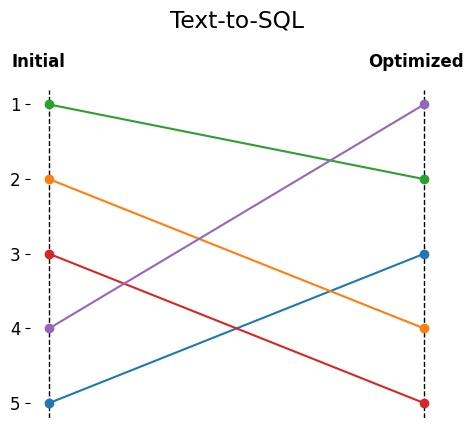} \hfill
    \includegraphics[width=0.195\linewidth]{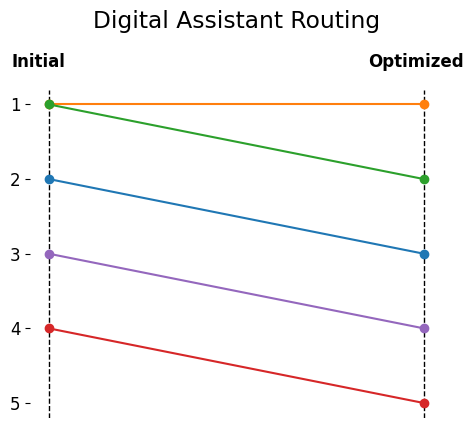} \hfill
    \includegraphics[width=0.195\linewidth]{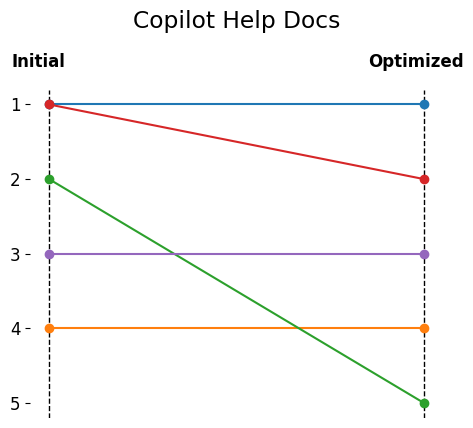} \hfill
    \includegraphics[width=0.195\linewidth]{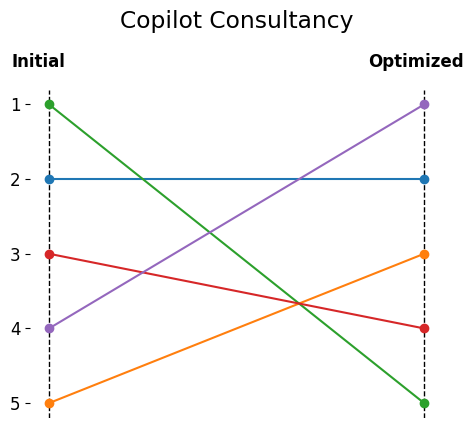} \hfill
    \includegraphics[width=0.195\linewidth]{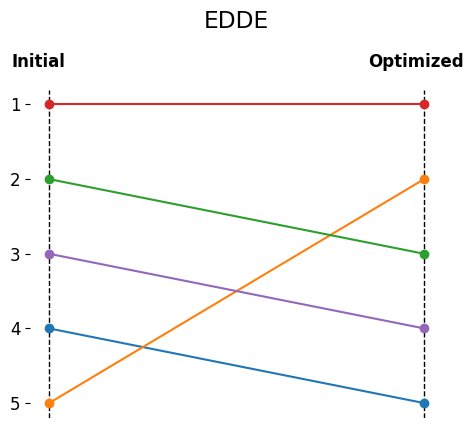}
    \caption{Rank changes across all models for datasets after instruction-only PO.\\\textbf{Top row: }Open-source datasets, \textbf{Bottom row:} Internal datasets.}
    \label{fig_2 - Ranking Changes, all datasets}
\end{figure*}

\begin{figure*}[htbp]
    \hspace{18.7mm}
    \includegraphics[width=0.195\linewidth]{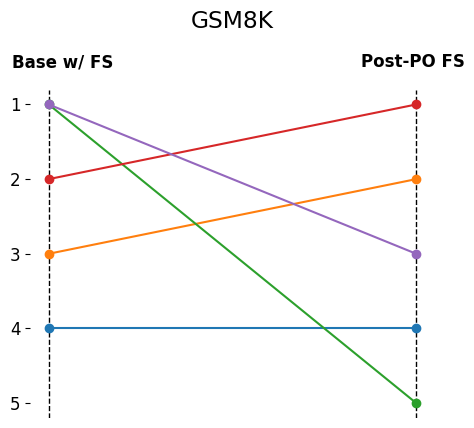} \hspace{15.9mm}
    \includegraphics[width=0.195\linewidth]{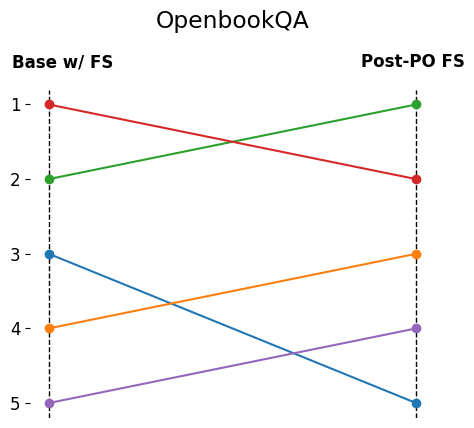} \hspace{15.9mm}
    \raisebox{0.03\height}{\includegraphics[width=0.265\linewidth]{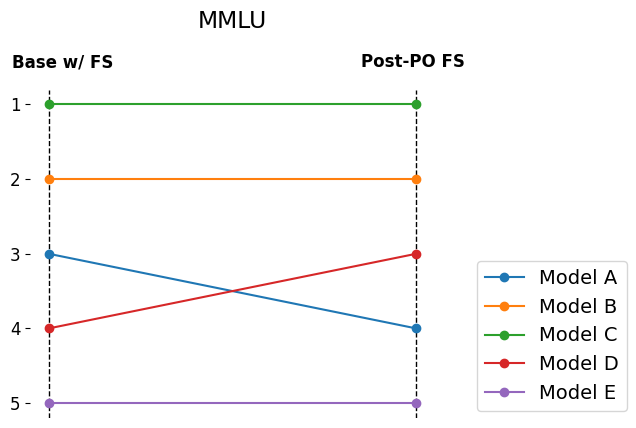}} 
    \hspace{11mm}
    \caption{Rank changes across all models for open-source datasets after instruction-with-exemplar PO.}
    \label{fig_2a - Ranking Changes, instruction+exemplar PO, open source datasets}
\end{figure*}

\section{Related Work} \label{sec_2:related works}
\subsection{Benchmarking Frameworks} \label{sec_2a:related works on benchmarking frameworks}
Earlier benchmarking frameworks are mainly a compilation of different tasks and metrics, packaged together with automated request APIs and clients, developed to allow simplified and automated evaluations of LLMs from multiple vendors on a variety of tasks from just one platform. Well-known examples include all variations of BigBench \cite{srivastava2023-bigbench}, and LM Eval Harness \cite{2024-eval-harness}. While these frameworks are well-regarded and very convenient, they are limited with regards to prompt-related features, for example, lacking any prompt engineering or built-in templates for the tasks. 

More recent frameworks have addressed these issues by including convenient features for prompt engineering or template creation \cite{saini-etal-2025-llm_evaluate-industry_focused_evaluation_tool_for_llms}. BigBio \cite{fries2022-bigbio} included a rudimentary interface that allowed users to engineer their own prompts before each evaluation run for all included biology-related tasks. HELM \cite{liang2023-helm-paper} improved this feature by allowing templates to be defined and saved, down to each subcategory of the wide taxonomy of tasks supported. Most recently, LLMeBench and Unitxt \cite{dalvi-etal-2024-llmebench, bandel2024-unitxt-flexible_shareable_reusable_benchmark} notably allow automated creation of prompt variations based on existing built-in task templates. 

However, none of these works includes automated per-model PO as part of their evaluation process or feature set. Our work aims to investigate whether PO should be a standard component in the pipeline of application-centric evaluations. 

\subsection{Prompt Optimization} \label{sec_2b:related works on prompt optimization}
Development of automated PO methods started due to well-documented observations that the quality of LLM generations is heavily dependent on the prompt quality and has preferences towards certain formatting \cite{zhang2022-tempera_paper-test_time_prompting_reinforcement, pryzant2023-automatic_prompt_optimization_gradient}, such as Claude models having preferences for XML tags \cite{anthropic-xml_preference}. The first category of PO methods focused only on optimizing the `instruction' portion of the prompt \cite{zhang2022-tempera_paper-test_time_prompting_reinforcement}. The second category focuses on the optimization of exemplars, based on the observations that exemplars have a greater influence on LLM performance.  \cite{yang2024-opro_paper-large_language_models_as_optimizers, yuksekgonul2024-textgrad_paper, wan2024-instructions_vs_exemplars-auto_prompt_optimization, liu2025_beyondpromptcontent_cfpo_paper}. 

Many types of methods have been explored, such as gradient descent \cite{yuksekgonul2024-textgrad_paper, pryzant2023-automatic_prompt_optimization_gradient}, reinforcement learning \cite{zhang2022-tempera_paper-test_time_prompting_reinforcement}, feedback-based methods \cite{pryzant2023-automatic_prompt_optimization_gradient}, and fine-grained Monte Carlo sampling \cite{liu2025_beyondpromptcontent_cfpo_paper}, showing the rapid development of PO methods in recent years. However, even with these developments, the integration of PO and optimized prompts as part of larger-scale evaluation frameworks has not yet been explored.

\section{Effects of Prompt Optimization on Evaluation Frameworks} \label{sec_4:prompt_opt effects on eval frameworks}

\subsection{Limitations of Static Prompts for Application-driven Development} \label{sec_3:problems with current eval frameworks}

In current benchmarking frameworks, an LLM model, $M$, generates test predictions $y_i^M$ following Equation (\ref{eq_1-cur_y_test}) by applying a single static prompt template $P_{static}$ for each sample $x_i^{test}$ of a task data set, consisting of $n$ samples. A metric function $J$ then scores these predictions against the corresponding set of ground truth answers $y_i^{test}$. The overall model score for task $S_M$ is the aggregation of individual scores; for simplicity, we restrict ourselves to the average, as shown in Equation (\ref{eq_2-score_eq}).

\begin{equation} \label{eq_1-cur_y_test}
y_i^M = M(P_{static}(x_i^{test})) 
\end{equation}
\begin{equation} \label{eq_2-score_eq}
S_M = \frac{1}{n}\Sigma_{i=0}^{n} J({y_i^{test}}, y_i^M)
\end{equation}
\begin{equation} \label{eq_3-obj_eq}
M^* = \arg \max_{M \in \mathcal{M} }{S_M}
\end{equation}
The goal of model evaluation is finding the model $M*$ with the highest score $S_M$ among all evaluated models $\mathcal{M}=[M_1, M_2, ...]$ , as shown in Equation (\ref{eq_3-obj_eq}). 

This use of $P_{static}$ makes $M$ the only optimizable variable to improve the score $S_M$. This approach is suitable for model-centric evaluation that assumes that the LLM is an interface that should be interoperable with any prompt. However, this does not fit the application-centric approach, where the input prompt $P$ is considered another optimizable variable to maximize the target objective.

\subsection{Brief Review of Prompt Optimization} \label{sec_4a:brief review of prompt optimization}

A complete prompt consists of three different components, as described in Equation (\ref{eq_a:prompt_definition}). First, the system prompt that dictates the `persona' of a model, followed by the task-specific instructions, $I$, describing the target task and recommended completion strategies. Optionally, this is followed with a few additional examples (`few-shot' exemplars), $E$, that further illustrate how tasks should be completed by the model. The final component is the main user query, $x$, to be solved by the model \cite{brown2020-gpt3_paper, alex2022-raft_few_shot_benchmark, zhang2024-sprig_improving_llm_system_prompt_optimization}. Note that in the domain of PO methods, $I$ usually refers to the combination of both the system and task-specific instruction components \cite{zhou2023-llms_are_human_level_prompt_engineers}. 
\begin{equation}  \label{eq_a:prompt_definition}
    P(x)=I+E+x
\end{equation}

The objective of any PO method, $F_{opt}$, as defined in Equation (\ref{eq_b:generatl_objective_of_po}), is to find the best $I$ and $E$ for a model $M$ that makes up the optimized prompt $P_M$ based on the existing base prompt $P_0 = I_0 + E_0$ and a set of training and validation samples, $x^{train}$ and $x^{valid}$. As discussed in Section \ref{sec_2b:related works on prompt optimization}, current PO methods can be categorized into those that focus only on $I^*_M$, and those that focus on $E^*_M$, or optimizing for both $I^*_M+E^*_M$ (instruction-with-exemplars) \cite{zhou2023instructionfollowingevaluationlargelanguage, zhou2023-llms_are_human_level_prompt_engineers, cheng2024-blackbox_prompt_optimization-align_llm_wo_modeltraining, wan2024-instructions_vs_exemplars-auto_prompt_optimization}. In this paper, experiments with the optimization of $I^*_M$ and $I^*_M + E^*_M$ will be explored.

\begin{equation} \label{eq_b:generatl_objective_of_po}
    \begin{aligned}
        P^*_M & = I^*_M + E^*_M \\
              & = F_{opt}(P_{0}, M, x^{train}, x^{valid})
    \end{aligned}
\end{equation}

\subsection{Effect of Prompt Optimization on Model Rankings in Evaluation Frameworks} \label{sec_4b:effects of prompt optimization on eval frameworks}
To address the limitations of $P_{static}$ mentioned in Section \ref{sec_3:problems with current eval frameworks}, an optimized prompt template per model, $P_M^*$, can first be obtained by following Equation \ref{eq_4-optimal_prompt}, that is, by applying a PO process $F_{opt}$ with the model $M$ to existing $P_{static}$ on the train data $x^{train}$. 

\begin{equation} \label{eq_4-optimal_prompt}
P^*_M = F_{opt}(P_{static}, M, x^{train}, x^{valid})
\end{equation}
\begin{equation} \label{eq_5-new_y_val}
{y^*}_i^M = M(P^*_M(x_i^{test})) 
\end{equation}
\begin{equation} \label{eq_6-new_score_eq}
S^*_M = \frac{1}{n}\Sigma_{i=0}^{n} J({y_i^{test}}, {y^*}_i^M)
\end{equation}
\begin{equation} \label{eq_7-new_obj}
M^* = \arg \max_{M \in \mathcal{M} }{S^*_M}
\end{equation}

The modified ${y^*}_i^M$ in Equation (\ref{eq_5-new_y_val}) substitutes $P_{static}$ in Equation (\ref{eq_1-cur_y_test}) with the optimized $P_M^*$, modifying the score and objective Equations (\ref{eq_2-score_eq}) and (\ref{eq_3-obj_eq}) of an evaluation framework, to become Equations (\ref{eq_6-new_score_eq}) and (\ref{eq_7-new_obj}), respectively. 
This score more accurately reflects the maximum possible performance of the model-prompt combination for the given task, affecting the final selection of $M^*$. 

\section{Experiments} \label{sec_5:experiment details}
The main objective of the experiments presented in this section is to verify the hypothesis that PO affects the choice of the best model for a given task. 

\subsection{{Model Details}}\label{sec_5a:details_of_models_tested}
{The experiments are carried out on five leading LLM models that are widely adopted in industry. For confidentiality reasons, we need to anonymize the model names. However, we can provide the following details about the models:} 


\begin{itemize}
    \item {Model A - closed-source multi-modal LLM, released in 2024. Claimed context length of 128K, with knowledge cutoff of October 2023.}
    \item {Model B - closed-source multi-modal LLM, released in 2024. Claimed context length of 1M+, with knowledge cutoff of May 2024.}
    \item {Model C - closed-source multi-modal LLM, released in 2024. Claimed context length of 200K, with knowledge cutoff of April 2024.}
    \item {Model D - open-weight text-only LLM, released in 2024. Instruction-tuned 8B parameter model, with context length of 128K.}
    \item {Model E - open-weight text-only LLM, released in 2024. Instruction-tuned 123B parameter model, and context length of 128K.}
\end{itemize}

\subsection{Prompt Optimization Setup} \label{sec_5b:prompt_opt_setup}
Two types of PO are implemented and tested: instruction-only and instruction-with-exemplar optimization. This adds another dimension to our experiments to highlight how much impact either type has on models' ranks. All optimization methods listed use GPT-4o \cite{openai2024-gpt4o_card} as the `critic' or optimizer model which provides iterative feedback on prompt selection.

Instruction-only optimization is implemented using the TextGrad framework \cite{yuksekgonul2024-textgrad_paper} with 8 training epochs, which take 3 optimization steps using batches of size 5. This means that per training epoch, at most 15 training examples are considered in the optimization, no matter the size of the training set (cf. Appendix \ref{app_a: full experimental details of datasets used}. Each step is followed by a validation step, in which the new proposed instruction is selected only if it can yield a higher score on the validation set than the previous one. Our implementation performs this validation step on the first 100 samples of the validation set. Instruction-only optimization is performed for all datasets listed in Section \ref{sec_5a:overview of dataset used}

For instruction-with-exemplar optimization, the `light' version of the MIPRO method is implemented using the DSPy framework \cite{opsahl_ong_etal_2024-mipro_paper, khattab2024-dspy_paper}. Optimization is performed for all open-source datasets listed in Section \ref{sec_5a:overview of dataset used}, with a maximum cap of 200 training and 300 validation samples. These number of samples are chosen because they are sufficiently large amount to obtain good exemplar optimization results but is still within the economical range of training samples encountered during practice. The results are compared to the `base' prompt with random examples chosen by HELM \cite{liang2023-helm-paper} to visualize the improvements made.

\subsection{Datasets} \label{sec_5a:overview of dataset used}
Two types of datasets are chosen for the experiments presented in this paper: open-source and internal datasets. This section will briefly describe the type of task represented by each dataset. Full details on experiment settings, such as split of each dataset, metrics, and ground truths used, are available in Appendix \ref{app_a: full experimental details of datasets used}.

\subsection*{Open-Source Datasets}
For open-source datasets, we utilize \textbf{GSM8K \cite{gsm8k_paper}}, \textbf{OpenbookQA \cite{openbookqa_paper}}, and \textbf{MMLU \cite{mmlu_paper}} due to their widespread adoption in multiple well-established frameworks and leaderboards \cite{huggingface-open-llm-leaderboard, liang2023-helm-paper, 2024-eval-harness}, representing generic problems used to evaluate LLMs.

\subsection*{Internal Datasets}
\begin{enumerate}
    \item \textbf{Digital Assistant Routing} is a dataset consisting of user queries to a digital assistant paired with labels that classify the type of request from the user. There are three category labels available: \texttt{TRANSACTIONAL}, \texttt{IR}, \texttt{ANALYTICS}
    \item \textbf{Copilot Help Docs} is a dataset created based on requests made to a business copilot chatbot. The LLMs task is to provide an answer to user queries about product documentation, based on context that is retrieved by the copilot. 
    \item \textbf{Copilot Consultancy}, is a dataset with a format similar to Copilot Help Docs. However, the questions and context are oriented to simulate users asking for information about company products, requiring the Copilot and the LLM to role-play as a consultant for the user. 
    \item \textbf{Text-To-SQL} is a dataset that consists of user requests containing data in JSON format that corresponds to a standard SQL database query. 
    \item \textbf{EDDE}, or Enterprise Document Data Extraction, is an information extraction dataset consisting of delivery note documents and ground truth of the extracted key-value pairs in JSON format.
\end{enumerate}
These datasets are chosen because they represent a diverse set of tasks, ranging from structured information extraction to open-ended QA problems such as consulting, and are derived from real industry use cases.

\section{Results and Discussions}
\begin{figure*}[htbp] 
    \includegraphics[width=0.49\linewidth]{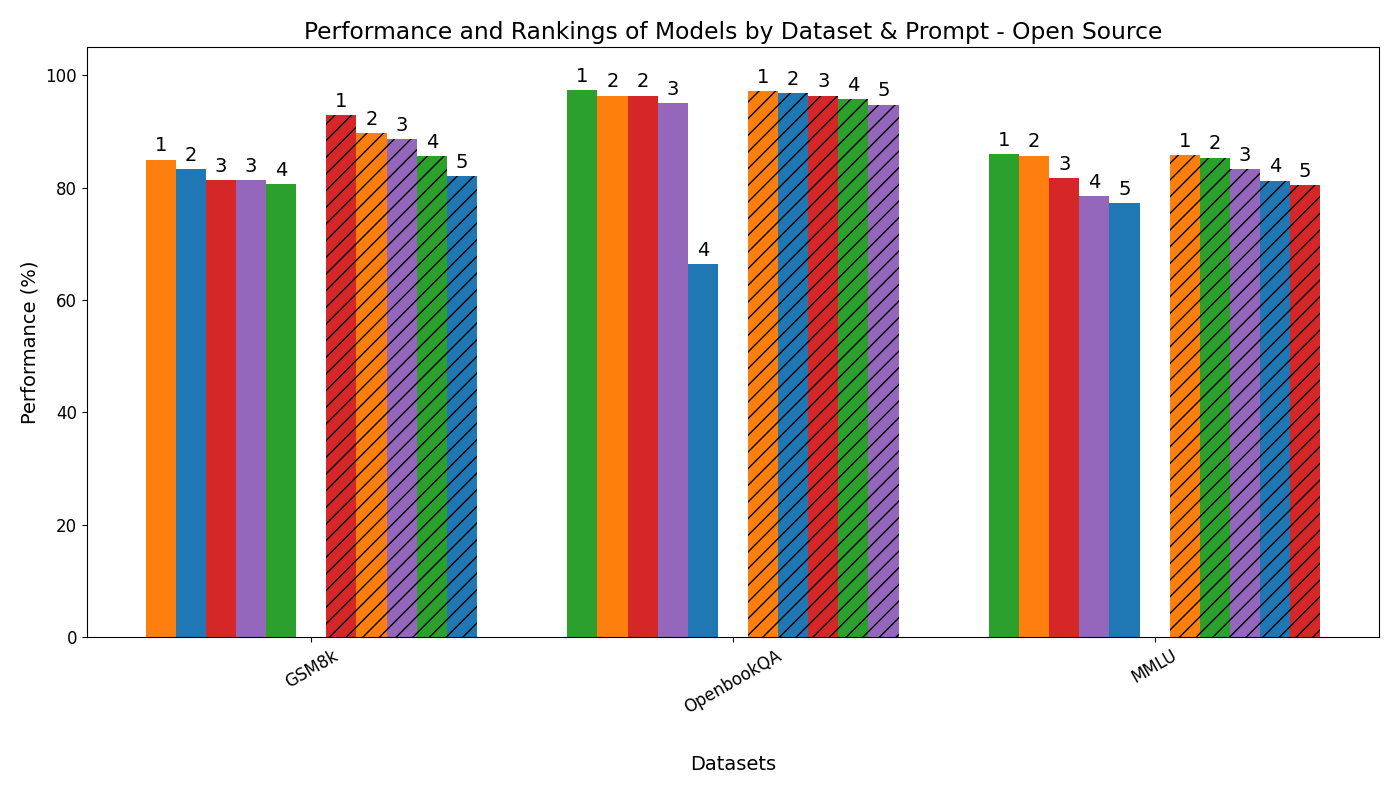} \hfill
    \includegraphics[width=0.49\linewidth]{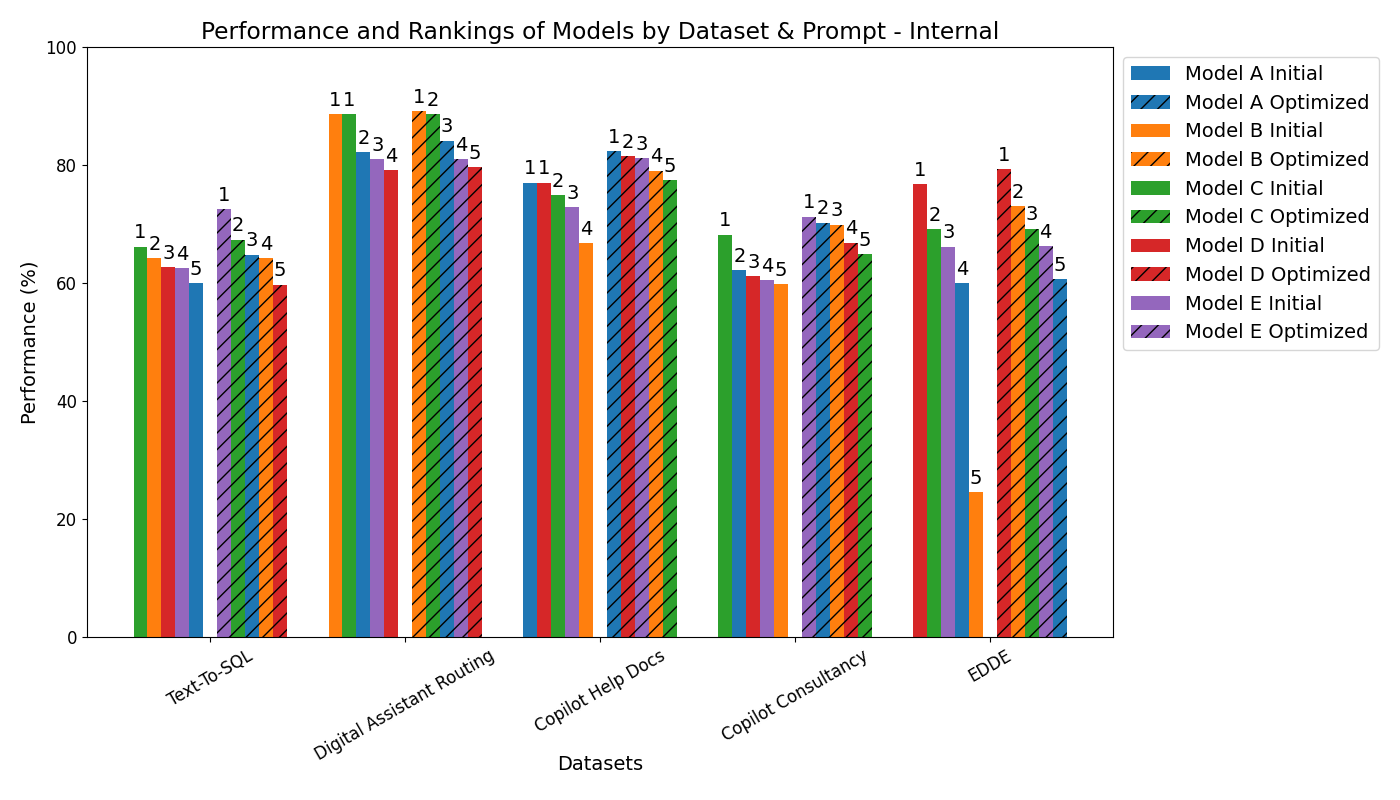}
    \caption{Performance and ranking values for all models, before and after instruction-only PO on tested datasets:\\ \textbf{Left:} Open-source, \textbf{Right:} Internal.}
    \label{fig_1 - Prompt Optimization results}
\end{figure*}

\begin{figure*}[htbp]
    \vspace{-10mm}
    \includegraphics[width=0.98\linewidth, center]{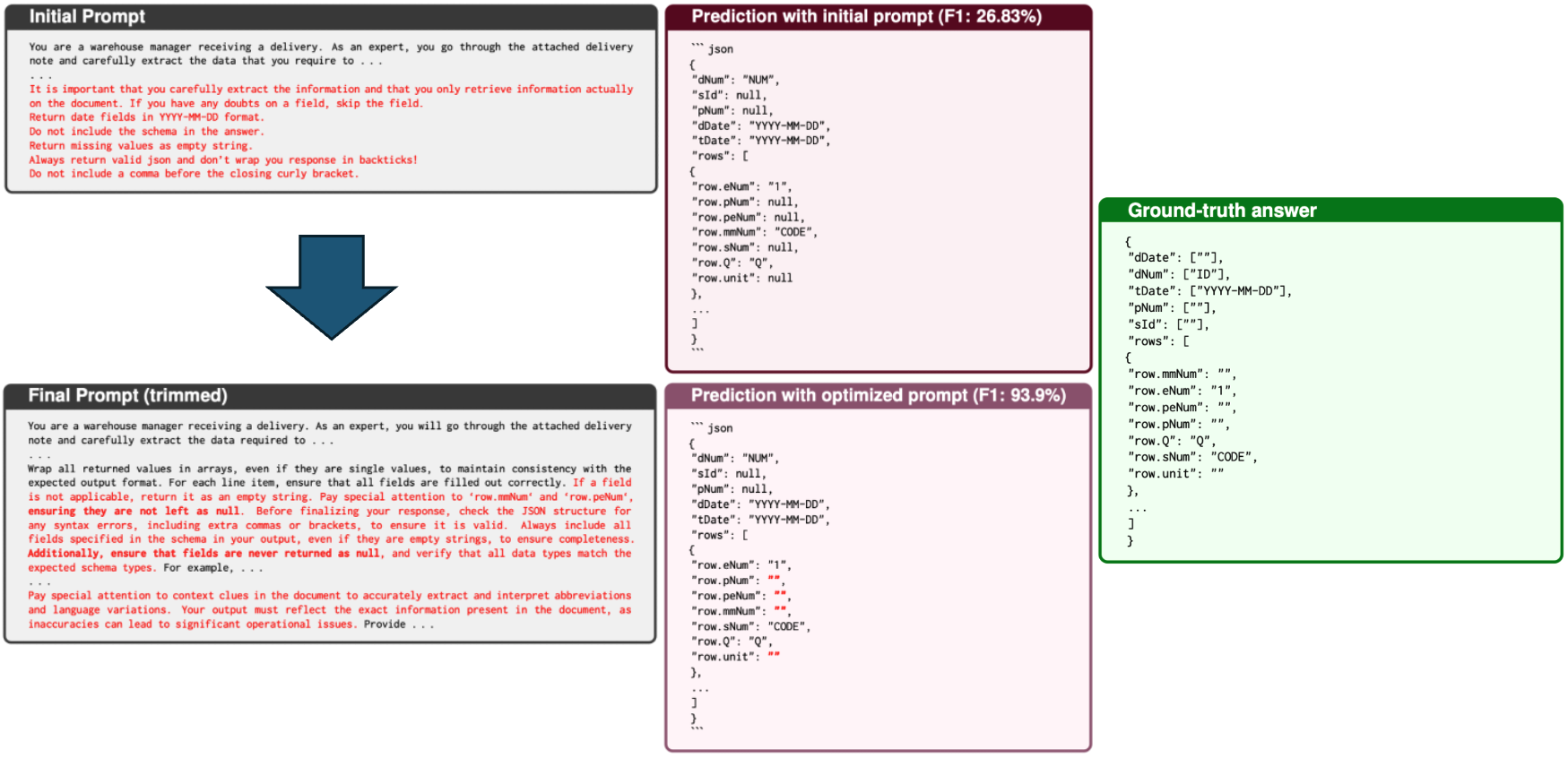}
    \caption{Example of instruction-following improvement after instruction-only PO on the EDDE dataset - Model B initially did not follow expected instructions and produced unintended `null' results. This is rectified using the optimized prompt, greatly improving the model's score for this sample question.}
    \label{fig_2-EX - Example of Instruction Following changes}
\end{figure*}
\begin{figure*}
    \includegraphics[width=0.95\linewidth, center]{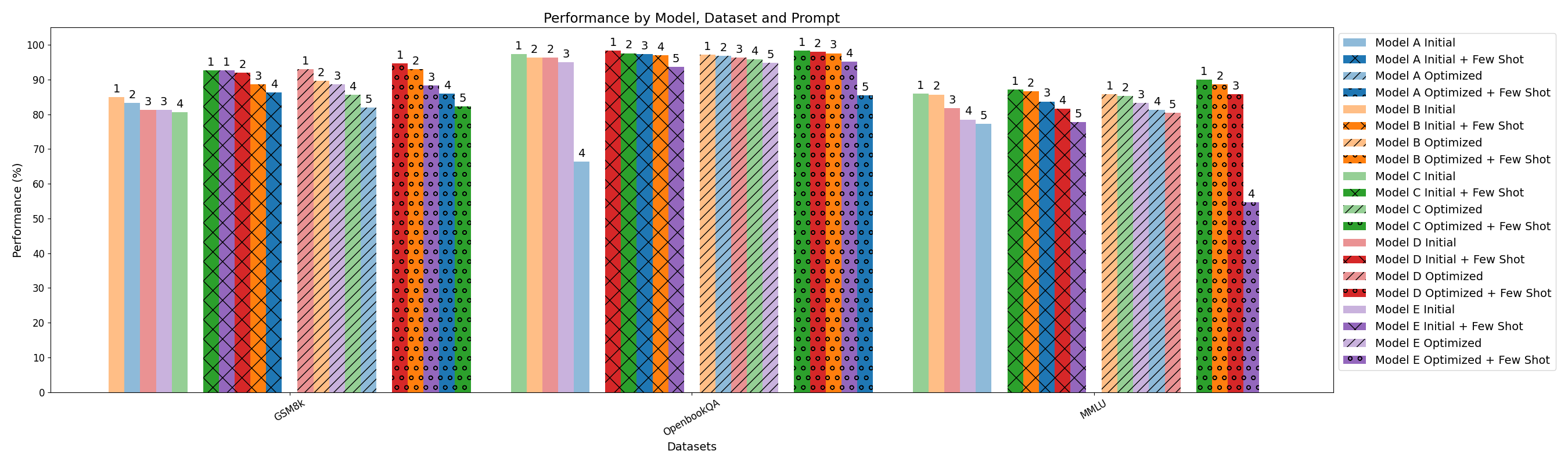}
    \caption{Performance and ranking changes after applying instruction-with-exemplar optimization.}
    \label{fig_4 - Results w/ Few-shot Exemplars}
\end{figure*}

\begin{figure}[htbp]
    \centering
    \includegraphics[width=0.98\linewidth, center]{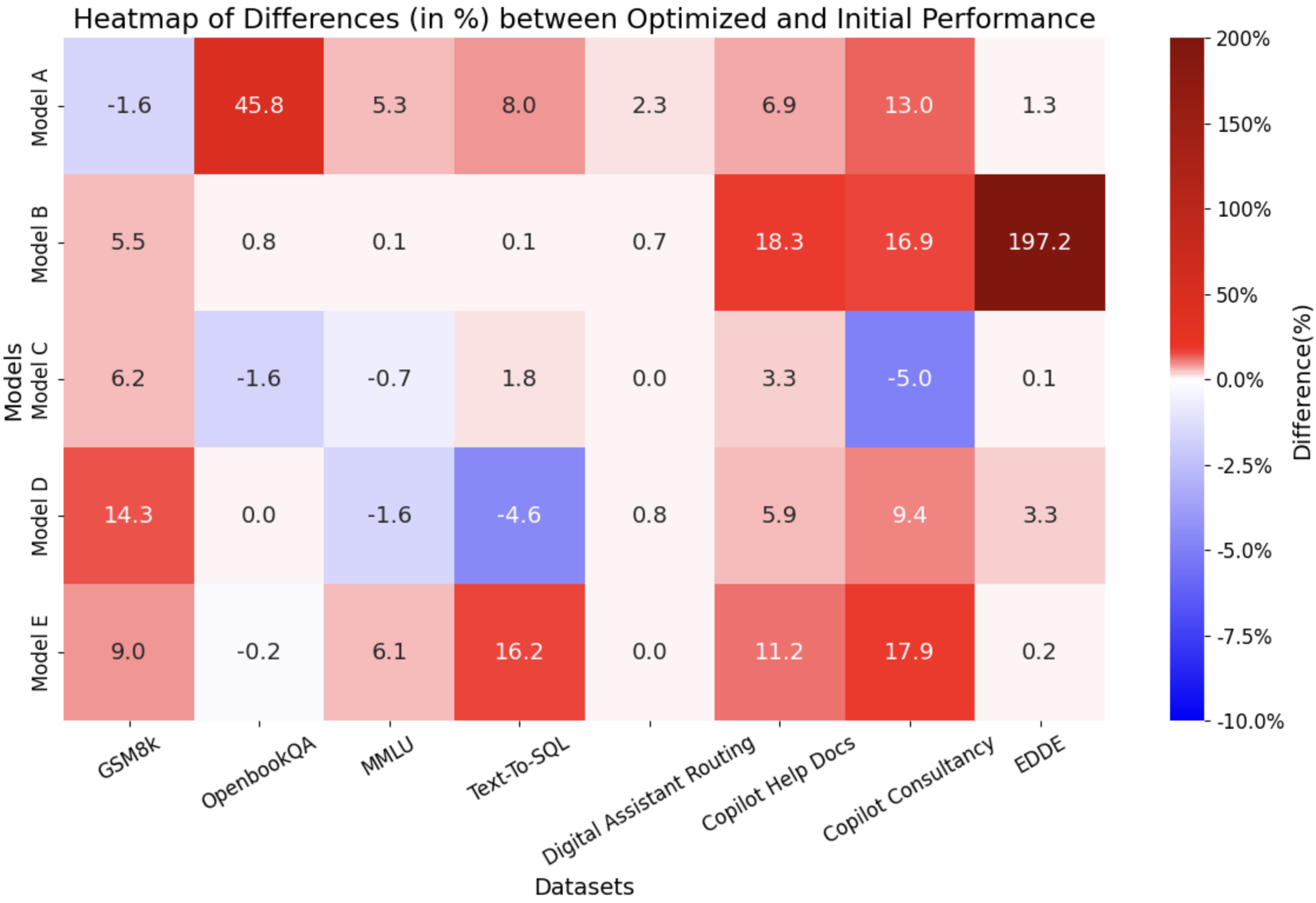}
    \caption{Heatmap of performance changes across all models, instruction-only PO.}
    \label{fig_3 - Model prompt sensitivity heatmap}
\end{figure}
The performance and rank changes of the models tested across all datasets before and after PO using instruction-only optimization can be seen in Fig. \ref{fig_2 - Ranking Changes, all datasets} and Fig. \ref{fig_1 - Prompt Optimization results}, while results and rank changes for instruction-with-exemplar optimization can be seen in Fig. \ref{fig_2a - Ranking Changes, instruction+exemplar PO, open source datasets} and Fig. \ref{fig_4 - Results w/ Few-shot Exemplars}, respectively. All numerical values of these reported performances are available in Appendix \ref{app_b:detailed experiment results}. The result of instruction-with-exemplar optimization for Model A on the MMLU dataset is omitted in Fig. \ref{fig_4 - Results w/ Few-shot Exemplars}, because the optimization method failed to produce any new optimized sets of instructions and exemplars. 

The results show that PO can affect the model leaderboard and conclusions for a task. For example, scores with baseline prompts would suggest Models B and C as the best models for GSM8K and MMLU. However, scores with instruction-only optimization show that Models D and B are the best models for these tasks, respectively. This rank-switching observation is also repeated for instruction-with-exemplar optimization, with Model D becoming the best GSM8K model post-PO, instead of Model B with only baseline prompt. Moreover, the example in Fig. \ref{fig_2-EX - Example of Instruction Following changes} shows that PO also improved instruction-following capabilities, which supports the increased model performances.

To better quantify these rank changes, we report Kendall's Tau \cite{kendalls_tau} between original and post-PO ranks for all datasets and PO methods, as seen in Tables \ref{tab_1-kendall_tau-inst_only} and \ref{tab_2-kendall_tau-inst_w_exemplars}. These measurements show that on average model rankings using PO prompts have positive but very weak correlation (< 0.5 Tau) to the rankings using default prompts. This means that PO greatly affects model rankings in general, further supporting the idea that PO should be integrated as a standard part of application-centric model evaluations.

\begin{table}[]
    \centering
    \begin{tabular}{|l|c|}
        \hline
        \textbf{Dataset} & \textbf{Kendall's $\tau$} \\
        \hline
        GSM8K                   & 0.10541 \\
        OpenbookQA              & -0.10541 \\
        MMLU                    & 0.40 \\
        Text-to-SQL             & 0.0  \\
        DA Routing              & 0.94868 \\
        Copilot Help Docs       & 0.52704 \\
        Copilot Consultancy     & -0.40 \\
        EDDE                    & 0.40 \\
        \hline
        \textbf{Mean $\tau$}       & \textbf{0.23446}\\
        \hline
    \end{tabular}
    \caption{Kendall's Tau values of rank changes using instruction-only optimization.}
    \label{tab_1-kendall_tau-inst_only}
\end{table}

\begin{table}[]
    \centering
    \begin{tabular}{|l|c|}
        \hline
        \textbf{Dataset} & \textbf{Kendall's $\tau$} \\
        \hline
        GSM8K                   & -0.10541 \\
        OpenbookQA              & 0.40 \\
        MMLU                    & 0.80 \\
        \hline
        \textbf{Mean $\tau$}       & \textbf{0.36486}\\
        \hline
    \end{tabular}
    \caption{Kendall's Tau values of rank changes using instruction-with-exemplar optimization.}
    \label{tab_2-kendall_tau-inst_w_exemplars}
\end{table}

Another observation is that all PO methods produced a new maximum performance score for all datasets, such as Copilot Help Docs having a 6.9\% higher maximum score through the instruction-only optimization performed for Model A. This shows that for application-centric evaluations, PO should be done as part of the evaluation to get the actual maximum performance for a model.

Next, as shown in the heatmap of Fig. \ref{fig_3 - Model prompt sensitivity heatmap}, all models have different sensitivities to prompt changes depending on the tasks. For example, all models seem to be relatively unaffected by PO for the Digital Assistant Routing task. However, Model B is notably very sensitive to PO for Copilot Help Docs, Copilot Consultancy, and EDDE tasks. These results also show that PO is more beneficial for complex and open-ended tasks, such as GSM8K, Copilot Help Docs, and Copilot Consultancy. Meanwhile, PO seems to not benefit models on tasks they are already very good at, such as OpenbookQA and Digital Assistant Routing. This means that the nature of the tasks evaluated should also be considered when observing the final model scores after applying PO.

Finally, while PO generally helps, there are instances where it decreases model performance instead. The possible causes for these occurrences are discussed in greater detail in Appendix \ref{app_d:negative_po_examples}, and may be rectifiable with other more sophisticated PO techniques, which will be explored in future work.

\section{Conclusions}
This paper highlighted the issues of unoptimized static prompts in current benchmarking frameworks. Then analysis and experimental results are presented across multiple models and datasets that highlight how PO significantly change the performance rankings of the models and affect the final model selections for the tasks tested. These results strongly support the recommendation that optimizing prompts should be incorporated as a standard procedure for any model evaluations in application-centric development. 


\section*{Limitations}
The results shown in this paper were produced using only two prompt optimization methods with one `critic' model. We did not {conduct repeated optimization tests to verify any standard deviation of the methods used. Next, we} did not consider the additional dimensions of the different prompt optimization methods and critic models available. Our work only considered the `black-box' usage of LLMs where weights are not fine-tuned. Additionally, this paper did not conduct `interoperability' experiments to see if the optimized prompts for one model are reusable to improve others. Our also work mainly considered `chat'-type models, and did not include tests with more recent `reasoning' models, such as Deepseek's R1 \cite{deepseekai2025-deepseek_r1}. Finally, we acknowledge that the model anonymizations imposed due to confidentiality requirements make the reported results difficult to reproduce.



\bibliography{custom.bib}

\appendix

\section{Dataset Details} \label{app_a: full experimental details of datasets used}
This section provides a more in-depth technical breakdown for the datasets used for the experiments, as mentioned in Section \ref{sec_5a:overview of dataset used}.
\subsection*{Open-Source}
\begin{enumerate}
    \item \textbf{GSM8K} - For our work, the dataset is split into train/validation/test of 200, 300, 300 respectively.
    Metric used is by extracting the last detected integer of a model's output string, which is then compared to the ground truth answer. The metric returns a final evaluation score of either one (match) or zero (no match) per sample, and the final reported score is the average of the results from all samples tested.
    
    \item \textbf{OpenbookQA} - Usually, this dataset requires the LLM to perform information retrieval (IR) from the provided facts list and use it to generate a final answer. However, for this paper a simplified version is used, skipping the IR step due to it being outside the scope of this article, and pairing the most relevant fact as context for each question. These changes simplify the dataset to simplify the dataset to use only a singular metric, in-line with the other chosen datasets. 
    
    Evaluations of model predictions are done by first extracting the answers via regular expression \cite{explanations_for_regular_expressions} that matches for the string “Answer:” followed by the actual one-letter answer. The extracted letter can then be compared against the ground truth using exact match metric. The ground truths consist of just a capitalized letter from ‘A’ to ‘D’ corresponding to one of the four available answers, producing an accuracy score. The final reported performance score for this dataset indicate the average accuracy score across all test samples. For this paper, 500 samples are used for testing and another 500 for validation steps in the training process. The remaining 4957 samples are used as training data.

    \item \textbf{MMLU} - For this paper, we have chosen five subjects from the list supported by MMLU: abstract algebra, econometrics, conceptual physics, machine learning, and professional medicine. These topic choices are based on their diversity covering a wide range of subjects. Additionally, the similar performance of GPT-3 across these topics, as reported in the original MMLU paper, also suggests similar dataset quality across these topics \cite{mmlu_paper, brown2020-gpt3_paper}. 
    Since the final format is a multiple-choice answer similar to OpenbookQA, the same regular expression-based metric and final performance score are also utilized. Commonly available train/validation/test for the 5 tasks are used and concatenated which results in a 25/91/833 split.
\end{enumerate}

\subsection*{Internal}
\begin{enumerate}
    \item \textbf{Digital Assistant Routing} - The evaluation of model predictions for this dataset is done by direct comparison to the ground truth, leading a score of zero or one. The final reported score is the average value of these scores. The train/validation/test split used for the experiment results shown is 735/157/158.
    \item \textbf{Copilot Help Docs} - To evaluate model predictions, a human-aligned satisfactory answer is provided as the ground truth. The LLM answer and the ground truth are compared using an LLM as a judge setup, with GPT-4o \cite{openai2024-gpt4o_card} utilized as the `judge' model. This setup uses a prompt that leads the judge LLM to rate the answer with a score from one to five, and a reasoning behind its rating. This judge rating is then linearly normalized to a final score between zero to one as the final metric score, to better align with the metric values used for other dataset. There are a total of 311 data samples available in this dataset, and the train/validation/test split of 150/100/61 is used for the experiments detailed in this paper.
    \item \textbf{Copilot Consultancy} - Due to the similar open-ended nature of the task, the same LLM as a judge setup for Copilot Help Docs is used for the evaluation metric. This dataset has 374 available samples, segregated into 200/100/74 split for train/validation/test.  
    \item \textbf{Text-to-SQL} - For evaluation metric, each predictions are scored by comparing how many fields and values (entries) in the predicted JSON string match with the entries of ground truth JSON. The final score reported for this task is the average precision of the JSON entries generated by the model. This dataset is rather small with only 56 available samples. For the results, all 56 are used as the test set. The training process uses a train set of 47 samples and a validation set of 7 samples.
    \item \textbf{EDDE} - Evaluating a prediction for EDDE works similarly to Text-to-SQL, however the final metric uses the F1 score of the predicted entries instead. The train/validation/test split used is 200/100/96.
\end{enumerate}

\section{Detailed Experiment Results} \label{app_b:detailed experiment results}
\label{sec:appendix}
Detailed numerical values for results shown in Figures \ref{fig_1 - Prompt Optimization results} and \ref{fig_4 - Results w/ Few-shot Exemplars} is available in Table \ref{tab_app_b:all experiment results}.

\begin{table*}
    \centering
    \begin{tabular}{lccccc}
        \hline
        \textbf{Dataset} & \textbf{Model} & \textbf{Initial} & \textbf{Initial w/ FS} & \textbf{Optimized} & \textbf{Optimized w/ FS} \\
        \hline
        \multirow{4}{*}{GSM8K}      & Model A & 83.33\% & 86.33\% & 82.00\% & 86.00\%\\
                                    & Model B & 85.00\% & 88.67\% & 89.67\% & 93.00\%\\
                                    & Model C & 80.67\% & 92.66\% & 85.67\% & 82.30\%\\
                                    & Model D & 81.33\% & 92.00\% & 93.00\% & 94.67\%\\
                                    & Model E & 81.33\% & 92.66\% & 88.67\% & 88.30\%\\
        \hline
        \multirow{4}{*}{OpenbookQA} & Model A & 66.40\% & 97.40\% & 96.80\% & 85.40\%\\
                                    & Model B & 96.40\% & 97.00\% & 97.20\% & 97.60\%\\
                                    & Model C & 97.40\% & 97.60\% & 95.80\% & 98.40\%\\
                                    & Model D & 96.40\% & 98.40\% & 96.40\% & 98.00\%\\
                                    & Model E & 95.00\% & 93.60\% & 94.80\% & 95.20\%\\
        \hline
        \multirow{4}{*}{MMLU}       & Model A & 77.19\% & 86.33\% & 82.00\% & 86.00\%\\
                                    & Model B & 85.71\% & 88.67\% & 89.67\% & 93.00\%\\
                                    & Model C & 85.95\% & 92.66\% & 85.67\% & 82.30\%\\
                                    & Model D & 81.75\% & 92.00\% & 93.00\% & 94.67\%\\
                                    & Model E & 78.51\% & 92.66\% & 88.67\% & 88.30\%\\
        \hline
        \multirow{4}{*}{Text-to-SQL}        & Model A & 60.05\% & N/A     & 64.84\% & N/A    \\
                                            & Model B & 64.23\% & N/A     & 64.30\% & N/A    \\
                                            & Model C & 66.21\% & N/A     & 67.38\% & N/A    \\
                                            & Model D & 62.66\% & N/A     & 59.75\% & N/A    \\
                                            & Model E & 62.49\% & N/A     & 72.59\% & N/A    \\
        \hline
        \multirow{4}{*}{DA Routing}         & Model A & 82.28\% & N/A     & 84.18\% & N/A    \\
                                            & Model B & 88.61\% & N/A     & 89.24\% & N/A    \\
                                            & Model C & 88.61\% & N/A     & 88.61\% & N/A    \\
                                            & Model D & 79.11\% & N/A     & 79.75\% & N/A    \\
                                            & Model E & 81.01\% & N/A     & 81.01\% & N/A    \\
        \hline
        \multirow{4}{*}{Copilot Help Docs}    & Model A & 77.05\% & N/A     & 82.38\% & N/A    \\
                                            & Model B & 66.85\% & N/A     & 79.10\% & N/A    \\
                                            & Model C & 75.00\% & N/A     & 77.46\% & N/A    \\
                                            & Model D & 77.05\% & N/A     & 81.56\% & N/A    \\
                                            & Model E & 72.95\% & N/A     & 81.15\% & N/A    \\
        \hline
        \multirow{4}{*}{Copilot Consultancy}  & Model A & 62.16\% & N/A     & 70.27\% & N/A    \\
                                            & Model B & 59.80\% & N/A     & 69.93\% & N/A    \\
                                            & Model C & 68.24\% & N/A     & 64.86\% & N/A    \\
                                            & Model D & 61.15\% & N/A     & 66.89\% & N/A    \\
                                            & Model E & 60.47\% & N/A     & 71.28\% & N/A    \\
        \hline
        \multirow{4}{*}{EDDE}                & Model A & 59.94\% & N/A     & 60.74\% & N/A    \\
                                            & Model B & 24.60\% & N/A     & 73.10\% & N/A    \\
                                            & Model C & 69.16\% & N/A     & 69.25\% & N/A    \\
                                            & Model D & 76.86\% & N/A     & 79.36\% & N/A    \\
                                            & Model E & 66.20\% & N/A     & 66.35\% & N/A    \\
        \hline           
    \end{tabular}
    \caption{All recorded results for the experiments conducted.}
    \label{tab_app_b:all experiment results}
\end{table*}

\section{Inference Samples}
The following is the list of sample questions, ground truth, predictions, and prompts demonstrating examples of improvements provided by PO throughout the conducted experiments. All important prompt and prediction changes between the initial and post-PO process, relative to the respective tasks, are highlighted in red color text. 

\subsection{Model D for GSM8K}

This sample shows how Model D's responses are improved by instruction-only PO for GSM8K. The response using the initial `base' prompt correctly shows the ground truth answer of 91 but also includes the number 10 in its full answer sentence. Since the metric function extracts the last detected number in the response string, 10 is registered as the model's answer which is deemed incorrect. The final prompt provided by instruction-only optimization improved the model's answer format to only include one number in the last sentence, turning the answer to be correct under the GSM8K's metric function.

This improvement is especially apparent when looking at the last sentence of the optimized prompt:
\begin{tcolorbox}[boxsep=1pt]
    \tcbfontsize{0.75} \texttt{Highlight the final answer prominently at the end of your response.}
\end{tcolorbox}
showing how the PO process is able to identify the issue of final answer formatting and tweak the instruction prompt to fix this issue for the GSM8K task.

\begin{tcolorbox}[fit basedim=10pt, colback=blue!5!white,colframe=blue!50!black, fonttitle=\sffamily\bfseries\small,fontupper=\sffamily,title=Sample Question]
    \tcbfontsize{0.65}\texttt{Tom plants 10 trees a year. Every year he also chops down 2 trees a year.  He starts with 50 trees.  After 10 years 30\% of the trees die.  How many trees does he have left?}
\end{tcolorbox}
\begin{tcolorbox}[fit basedim=10pt, colback=green!5!white,colframe=green!50!black, fonttitle=\sffamily\bfseries\small, fontupper=\sffamily, title=ground truth answer]
    \tcbfontsize{0.65}\texttt{91}
\end{tcolorbox}

\begin{breakableTcolorboxInitPredict}{Prediction with the initial prompt (Wrong Answer)}
    
    \tcbfontsize{0.65} \texttt{To solve this problem, let's break it down step by step.\\
    1. Tom starts with 50 trees.\\
    2. Every year, he plants 10 trees and chops down 2 trees. So, the net gain of trees per year is 10 - 2 = 8 trees.\\
    3. Since this process repeats for 10 years, the total net gain of trees in 10 years is 8 * 10 = 80 trees.\\
    4. After 10 years, the total number of trees Tom has is the initial number of trees (50) plus the net gain of trees (80), which is 50 + 80 = 130 trees.\\
    5. However, after 10 years, 30\% of the trees die. To find the number of trees that die, we calculate 30\% of 130, which is 0.3 * 130 = 39 trees.\\
    6. Finally, to find the number of trees Tom has left, we subtract the number of trees that died (39) from the total number of trees he had after 10 years (130), which is 130 - 39 = 91 trees.\\
    \textcolor{red}{Therefore, Tom has 91 trees left after 10 years.}
    }
\end{breakableTcolorboxInitPredict}
\begin{breakableTcolorboxOptPredict}{Prediction with optimized prompt \\(Correct Answer)}
    \tcbfontsize{0.65}\texttt{To find the number of trees Tom has left after 10 years, we need to follow these steps:\\
    1. Calculate the net number of trees Tom plants each year: 10 trees (planted) - 2 trees (chopped down) = 8 trees per year.\\
    2. Calculate the total number of trees Tom adds over 10 years: 8 trees/year * 10 years = 80 trees.\\
    3. Calculate the total number of trees Tom has after 10 years, before any die: Initial trees + trees added = 50 trees + 80 trees = 130 trees.\\
    4. Calculate the number of trees that die after 10 years: 30\% of 130 trees = 0.3 * 130 = 39 trees.\\
    5. Calculate the final number of trees Tom has left: Total trees - trees that died = 130 trees - 39 trees = 91 trees.\\
    \textcolor{red}{**The final answer is: 91**}
    }
\end{breakableTcolorboxOptPredict}

\begin{breakableTcolorbox}{Initial Prompt}
\tcbfontsize{0.65}\texttt{As an expert problem solver solve step by step the following mathematical questions.}
\end{breakableTcolorbox}

\begin{breakableTcolorbox}{Final Prompt} 
\tcbfontsize{0.65}\texttt{Provide a clear and concise solution to the following mathematical question, detailing each step of the calculation. Ensure that each step is clearly articulated, use precise language to convey the calculations and the final result, and avoid unnecessary redundancy. \textcolor{red}{Highlight the final answer prominently at the end of your response.}}
\end{breakableTcolorbox}

\subsection{Model E for Text-To-SQL}
In this sample, Model E generated a response that had unnecessary explanation strings in addition to the valid JSON answer using the initial prompt. These additional strings caused an error with the metric function and returned a 0\% score. The optimized prompt rectified this problem, forcing Model E to generate only the JSON response and get a proper metric score. 
\begin{tcolorbox}[fit basedim=10pt, fonttitle=\sffamily\bfseries\small, fontupper=\sffamily, colback=blue!5!white,colframe=blue!50!black,title=Sample Question]
\tcbfontsize{0.65} \texttt{Count Product and Location by Store}
\end{tcolorbox}
\begin{breakableTcolorboxGTAnswer} {ground truth answer}
\tcbfontsize{0.65} \texttt{\{\\
"dimensions": [\\
\{\\
"entity": "Product",\\
"locale": "en-US"\\
\},\\
\{\\ 
"entity": "Location",\\
"locale": "en-US"\\
\}\\
],\\
"measures": [\\
\{\\
"entity": "Product",\\ 
"calculation": \{\\ 
"aggregation": \{\\ 
"method": "Dimension Count",\\
"base\_dimension": "Store"\\
\}\\
\}\\
\},\\
\{\\
"entity": "Location",\\
"calculation": \{\\
"aggregation": \{\\
"method": "Dimension Count",\\
"base\_dimension": "Store"\\
\}\\
\}\\
\}\\
],\\
"filters": []\\
\}}
\end{breakableTcolorboxGTAnswer}
\begin{breakableTcolorboxInitPredict}{Prediction with initial prompt \\(0\% - Error thrown)}
\tcbfontsize{0.65} \texttt{ \textcolor{red}{To analyze the question "Count Product and Location by Store" using Named Entity Recognition (NER) and categorize the entities into dimensions, measures, and filters for data analysis, we can break down the question as follows:\\
\\
1. **Dimensions**: These are the categorical data that describe the characteristics or attributes of the data.\\
2. **Measures**: These are the numerical data that quantify the dimensions and can be aggregated.\\
3. **Filters**: These are any conditions or constraints applied to the data.\\
\\
Given the question, we can identify the entities as follows:\\
\\
- **Dimensions**: "Store", "Product", "Location"\\
- **Measures**: "Count"\\
- **Filters**: None\\
\\
The calculation subtype for the measure "Count" is a simple count aggregation.\\
\\
Here is the JSON format for the identified entities:\\
\\}}
\textasciigrave\textasciigrave\textasciigrave\ \texttt{json\\
\{\\
  "dimensions": [\\
    \{"entity": "Store", "locale": "en-US"\},\\
    \{"entity": "Product", "locale": "en-US"\},\\
    \{"entity": "Location", "locale": "en-US"\}\\
  ],\\
  "measures": [\\
    \{"entity": "Count", "calculation": \{"aggregation": \{"method": "Count", "base\_dimension": "Store"\}\}\}\\
  ],\\
  "filters": []\\
\}\\}
\textasciigrave\textasciigrave\textasciigrave\ \\
\texttt{\textcolor{red}{This JSON structure categorizes the entities into dimensions, measures, and filters, and specifies the calculation subtype and base dimension for the measure}}
\end{breakableTcolorboxInitPredict}

\begin{breakableTcolorboxOptPredict}{Prediction with optimized prompt (60\%)}
\tcbfontsize{0.65}\textasciigrave\textasciigrave\textasciigrave\ \texttt{json\\
\{\\
  "dimensions": [\\
    \{"entity": "Store", "locale": "en-US"\},\\
    \{"entity": "Location", "locale": "en-US"\}\\
  ],\\
  "measures": [\\
    \{"entity": "Product", "calculation": \{"aggregation": \{"method": "Count", "base\_dimension": "Store"\}\}\}\\
  ],\\
  "filters": []\\
\}\\}
\textasciigrave\textasciigrave\textasciigrave\ 
\end{breakableTcolorboxOptPredict}
\begin{breakableTcolorbox}{Initial Prompt} 
\tcbfontsize{0.65}\texttt{I have an analytic question and I need to identify the entities within it using Named Entity Recognition (NER).\\
\\
Instruction: In the context of multidimensional data analysis, dimensions refer to categorical data that describe the characteristics or attributes of the data, while measures refer to numerical data that quantify the dimensions and can be aggregated. Calculations, such as averages, need to be based on a specific dimension to provide meaningful context for the aggregation. Please identify the entities in the following question and categorize them into dimension, measure, and filter for data analysis. For measures, also specify the calculation subtype and the base dimension if applicable for numeric aggregation. \textcolor{red}{Provide the results in JSON format.}\\
\\
Question: Show me the Average Gross Margin by Time\\
Answer:\\
\{"dimensions": [\{"entity": "Time", "locale": "en-US"\}], "measures": [\{"entity": "Gross Margin", "calculation": \{"aggregation": \{"method": "Average", "base\_dimension": "Time"\}\}\}], "filters": []\}\\
\\
Question: Show me the Average Gross Margin by Date\\
Answer:\\
\{"dimensions": [\{"entity": "Date", "locale": "en-US"\}], "measures": [\{"entity": "Gross Margin", "calculation": \{"aggregation": \{"method": "Average", "base\_dimension": "Date"\}\}\}], "filters": []\}\\
\\
Question: Show me the Average Gross Margin over Date\\
Answer:\\
\{"dimensions": [\{"entity": "Date", "locale": "en-US"\}], "measures": [\{"entity": "Gross Margin", "calculation": \{"aggregation": \{"method": "Average", "base\_dimension": "Date"\}\}\}], "filters": []\}\\
\\
Question: Show me the Average Gross Margin over Date using Sales Manager\\
Answer:\\
\{"dimensions": [\{"entity": "Date", "locale": "en-US"\}, \{"entity": "Sales Manager", "locale": "en-US"\}], "measures": [\{"entity": "Gross Margin", "calculation": \{"aggregation": \{"method": "Average", "base\_dimension": "Date"\}\}\}], "filters": []\}\\
\\
Question: Show me the Average Gross Margin by Sales Manager over Date\\
Answer:\\
\{"dimensions": [\{"entity": "Sales Manager", "locale": "en-US"\}, \{"entity": "Date", "locale": "en-US"\}], "measures": [\{"entity": "Gross Margin", "calculation": \{"aggregation": \{"method": "Average", "base\_dimension": "Sales Manager"\}\}\}], "filters": []\}\\}
\end{breakableTcolorbox}

\begin{breakableTcolorbox}{Final Prompt}
\tcbfontsize{0.65} \texttt{I have an analytic question and I need to identify the entities within it using Named Entity Recognition (NER).\\
\\
Instruction: In the context of multidimensional data analysis, dimensions are categorical data that describe the characteristics or attributes of the data. Measures are numerical data that quantify the dimensions and can be aggregated. Entities are typically nouns or noun phrases that represent real-world objects or concepts. Calculations, such as averages, sums, counts, max, min, etc., are verbs or verb phrases that represent mathematical operations and need to be based on a specific dimension to provide meaningful context for the aggregation. \\
\\
Please identify the entities in the following question and categorize them into dimension, measure, and filter for data analysis. \\
\\
- For measures, always specify the calculation subtype and the base dimension if applicable for numeric aggregation. This is a required field for all measures. \\
\textcolor{red}{- Provide the results in a structured JSON format, ensuring all necessary fields are included in the output, such as 'entity', 'locale', 'Dimension Count', 'base\_dimension', and 'filters'. \\
- The entities should be capitalized and the response should not include any additional unstructured text.} \\
\\
The JSON object should contain separate arrays for dimensions, measures, and filters, and each array should contain objects with specific fields. \\
\\
- For filters, identify the entity, the operator (like '<', '>', '=', etc.), and the specific value or range that is being filtered on. \\
- The 'entity' field should always be in lowercase. \\
\\
The order of dimensions in the 'dimensions' array is important and should be accurately predicted. The number of dimensions in the 'dimensions' array should match the number of dimensions in the input question. If the input question includes filters, they should be accurately predicted and included in the 'filters' field.\\
\\
Use the context of the input question to generate a more accurate output, especially when predicting the order of dimensions, the number of dimensions, and the presence of filters. Aim to generate correct SQL queries for a wide range of inputs, and strive for robustness in your output. \\
\\
Ensure all entities and dimensions mentioned in the question are included in the response. Missing entities or dimensions will result in an incomplete response and a lower evaluation score. Follow the exact structure and formatting of the JSON object as shown in the examples. Any discrepancies in structure or formatting will result in a lower evaluation score. Avoid using placeholders in the response. The response should include specific entities or dimensions based on the input question. \\
\\
If an error is detected in the response, generate a new, corrected response. \\
\\
Here are some diverse examples:\\
\\
Question: Show me the Average Gross Margin by Time\\
Answer:\\
\{"dimensions": [\{"entity": "Time", "locale": "en-US"\}], "measures": [\{"entity": "Gross Margin", "calculation": \{"aggregation": \{"method": "Average", "base\_dimension": "Time"\}\}\}], "filters": []\}\\
\dots\\
Question: Show me the Total Sales by Region\\
Answer:\\
\{"dimensions": [\{"entity": "Region", "locale": "en-US"\}], "measures": [\{"entity": "Sales", "calculation": \{"aggregation": \{"method": "Total", "base\_dimension": "Region"\}\}\}], "filters": []\}\\
\\
Question: Show me the Count of Products sold over Date using Sales Manager\\
Answer:\\
\{"dimensions": [\{"entity": "Date", "locale": "en-US"\}, \{"entity": "Sales Manager", "locale": "en-US"\}], "measures": [\{"entity": "Products", "calculation": \{"aggregation": \{"method": "Count", "base\_dimension": "Date"\}\}\}], "filters": []\}\\
\\
Question: Show me the Max Revenue by Sales Manager over Date with a filter on Region\\
Answer:\\
\{"dimensions": [\{"entity": "Sales Manager", "locale": "en-US"\}, \{"entity": "Date", "locale": "en-US"\}], "measures": [\{"entity": "Revenue", "calculation": \{"aggregation": \{"method": "Max", "base\_dimension": "Sales Manager"\}\}\}], "filters": [\{"entity": "Region", "operator": "=", "value": "North America"\}]\}\\
\\
Question: What's the Median Gross Margin?\\
Answer:\\
\{"dimensions": [], "measures": [\{"entity": "Gross Margin", "calculation": \{"aggregation": \{"method": "Median"\}\}\}], "filters": []\}\\
\\
Question: What's the total count of products?\\
Answer:\\
\{"dimensions": [], "measures": [\{"entity": "Products", "calculation": \{"aggregation": \{"method": "Count"\}\}\}], "filters": []\} </LM\_SYSTEM\_PROMPT>}
\end{breakableTcolorbox}

\subsection{Model E for MMLU} \label{app_c3:mmlu_modele_instruction_only_po}
Model E provided the wrong answer using the initial prompt, due to the lack of careful step-by-step analysis of each statement provided in the question. This is rectified by the optimized prompt which added more thorough instruction paragraphs on how to parse the statements and provide the appropriate explanations for the answer. 
\begin{breakableTcolorboxSampleQuestion}{Sample Question}
\tcbfontsize{0.65}\texttt{Question: Statement 1 | A factor group of a non-Abelian group is non-Abelian. Statement 2 | If K is a normal subgroup of H and H is a normal subgroup of G, then K is a normal subgroup of G.\\
Choices:\\
A. True, True\\
B. False, False\\
C. True, False\\
D. False, True}
\end{breakableTcolorboxSampleQuestion}
\begin{tcolorbox}[fit basedim=10pt, fonttitle=\sffamily\bfseries\small, fontupper=\sffamily, colback=green!5!white,colframe=green!50!black,title=ground truth answer]
\tcbfontsize{0.65}\texttt{B}    
\end{tcolorbox}

\begin{tcolorbox}[fit basedim=10pt, fonttitle=\sffamily\bfseries\small, fontupper=\sffamily, colback=purple!5!white,colframe=purple!50!black,title=Prediction with initial prompt (Wrong Answer)]
\tcbfontsize{0.65}\texttt{Answer: C\\
\\
Explanation:\\
- Statement 1 is true. A factor group (or quotient group) of a non-Abelian group can indeed be non-Abelian.\\
- Statement 2 is false. If K is a normal subgroup of H and H is a normal subgroup of G, it does not necessarily mean that K is a normal subgroup of G. Normality is not transitive in general\\}
\end{tcolorbox}

\begin{breakableTcolorboxOptPredict} {Prediction with the optimized prompt (Correct Answer)}
\tcbfontsize{0.65}\texttt{\textcolor{red}{\#\#\# Analysis and Explanation\\
\\
\#\#\#\# Statement 1: A factor group of a non-Abelian group is non-Abelian.\\
\\
To analyze this statement, we need to understand the properties of factor groups (also known as quotient groups) and non-Abelian groups.\\
- **Non-Abelian Group**: A group \( G \) is non-Abelian if there exist elements \( a, b \in G \) such that \( ab \neq ba \).
- **Factor Group**: Given a group \( G \) and a normal subgroup \( N \), the factor group \( G/N \) consists of the cosets of \( N \) in \( G \).\\
\\
A factor group \( G/N \) can be Abelian even if \( G \) is non-Abelian. For example, consider the group of symmetries of a square, which is non-Abelian. If we take the normal subgroup consisting of the identity and a 180-degree rotation, the factor group is Abelian.\\
\\
Therefore, Statement 1 is **False**.\\}
\\
\textcolor{red}{\#\#\#\# Statement 2: If K is a normal subgroup of H and H is a} \textcolor{red}{normal subgroup of G, then K is a normal subgroup of G.\\
To analyze this statement, we need to understand the concept of normal subgroups.\\
- **Normal Subgroup**: A subgroup \( N \) of a group \( G \) is normal if \( gNg^{-1} = N \) for all \( g \in G \).\\
If \( K \) is a normal subgroup of \( H \) and \( H \) is a normal subgroup of \( G \), it does not necessarily mean that \( K \) is a normal subgroup of \( G \). For example, consider the quaternion group \( Q_8 \) which has a normal subgroup \( H \) of order 4, and \( H \) has a normal subgroup \( K \) of order 2. However, \( K \) is not normal in \( Q_8 \).\\
Therefore, Statement 2 is **False**.\\
\#\#\# Conclusion\\
Based on the analysis above, the correct answer is:\\}
\\
**Answer: B**\\
\\
\#\#\# Explanation Summary\\
- **Statement 1**: A factor group of a non-Abelian group can be Abelian, so the statement is false.\\
- **Statement 2**: If \( K \) is a normal subgroup of \( H \) and \( H \) is a normal subgroup of \( G \), \( K \) is not necessarily a normal subgroup of \( G \), so the statement is false.\\
Thus, the correct choice is **B. False, False**.}
\end{breakableTcolorboxOptPredict}

\begin{breakableTcolorbox}{Initial Prompt}
\tcbfontsize{0.65}\texttt{The following are multiple choice questions (with answers) about abstract\_algebra, conceptual\_physics, econometrics, machine\_learning and professional\_medicine. Provide your answer in the following format: "Answer: X", where X is a letter from A to D.}
\end{breakableTcolorbox}

\begin{breakableTcolorbox}{Final Prompt}
\tcbfontsize{0.65}\texttt{You will answer multiple choice questions designed to assess knowledge in the following areas: \\
- Abstract Algebra \\
- Conceptual Physics \\
- Econometrics \\
- Machine Learning\\ 
- Professional Medicine\\ 
For each question, present your answer first in the format "Answer: X" (where X is a letter from A to D). Ensure that your answer is clearly separated from the explanation to enhance readability. \\
\\
\textcolor{red}{Your explanation should provide a thorough rationale that connects your choice to the key concepts involved, including a detailed analysis of each answer option. Clearly justify your selected answer by linking it to the patient's symptoms and relevant medical context, while also explaining why the other options do not apply. \\
\\
Structure your response logically: begin by summarizing the key findings or symptoms presented in the question, then analyze each statement or option in a concise manner, and conclude with your answer. Aim for clarity and conciseness in your explanations, avoiding redundancy and unnecessary elaboration.\\
\\
Whenever possible, include relevant examples or analogies to illustrate complex concepts and enhance understanding. Use precise medical terminology to convey professionalism and depth of knowledge.\\
\\
Finally, self-evaluate your response for clarity, relevance, and adherence to the required format before finalizing your answer. Ensure that your statements are free of ambiguity and fully informative, reflecting a comprehensive grasp of the relevant theories and principles.\\}}    
\end{breakableTcolorbox}

\subsection{Model A for Digital Assistant Routing}
In this sample Model A initially came to the wrong answer using the initial prompt, before generating the correct response when using the optimized prompt. The most obvious difference here, other than the modified definitions of the categories, are the modified strategies in the optimal prompt for any potentially ambiguous questions, likely making Model A to re-assess its `thinking` process before arriving at its final answer.
\begin{tcolorbox}[fit basedim=10pt, fonttitle=\sffamily\bfseries\small, fontupper=\sffamily, colback=blue!5!white,colframe=blue!50!black,title=Sample Question]
\tcbfontsize{0.65}\texttt{How can I create credit and debit memo requests?}
\end{tcolorbox}

\begin{tcolorbox}[fit basedim=10pt, fonttitle=\sffamily\bfseries\small, fontupper=\sffamily, colback=green!5!white,colframe=green!50!black,title=ground truth answer]
\tcbfontsize{0.65}\texttt{IR}
\end{tcolorbox}

\begin{tcolorbox}[fit basedim=10pt, fonttitle=\sffamily\bfseries\small, fontupper=\sffamily, colback=purple!5!white,colframe=purple!50!black,title=Prediction with initial prompt]
\tcbfontsize{0.65}\texttt{TRANSACTIONAL}
\end{tcolorbox}
\begin{tcolorbox}[fit basedim=10pt, fonttitle=\sffamily\bfseries\small, fontupper=\sffamily, colback=magenta!5!white,colframe=magenta!50!black,title=Prediction with optimized prompt]
\tcbfontsize{0.65}\texttt{IR}
\end{tcolorbox}

\begin{breakableTcolorbox}{Initial Prompt}
    \tcbfontsize{0.65}\texttt{Your task is to classify the user query into one of the three query-type categories:\\
        - TRANSACTIONAL\\
        - IR\\
        - ANALYTICS\\
        \\
        TRANSACTIONAL: Transactional queries are also referred to as action queries. These queries are aimed at accomplishing personalized business-processes related task or action for the user. Types of actions that transactional queries perform are: create, add, get, update, delete, cancel, authorize, and approve. The tasks usually require special user permissions and access to backend systems. Transactional queries differ from IR queries in that transactional queries are individualized and typically require knowledge of the user's employee ID and authorized access to employee information systems in order to provide a relevant and user-specific answer. IR queries, on the other hand, can be answered from general company documentation and apply broadly according to company policies.\\
        \\
        IR: Information Retrieval (IR) queries seek answers to fact-finding questions regarding information that can be found in policy documents, user guides, support articles, learning content, or public content. Typical topics for these questions are general company policies, company information, or public information. This queries differ from transactional queries in that IR queries might ask general employee information-seeking questions regarding a work-related task, but transactional queries ask for an action to be performed that requires user-specific employee information and permissions.\\
        \\
        ANALYTICS: Analytics queries are natural language search-based data queries to our company's cloud analytics. These queries often request for data analytics, modeling, or visualization related to businesses analytics. These queries often resemble SQL and Hana-based queries. Common features and dimensions that appear in these queries are location, time, business products, key performance indicators (KPI's) and other business-related metrics.\\
        \\
        \textcolor{red}{Respond with only the category name in uppercase, without any additional text or punctuation.}\\
        \\
        Here are several examples of the user queries classifications:\\
        \\
        "Query": "show me calendar years in coach name point sort point \& week descend limiting 83 ok"\\
        "Classification": "ANALYTICS"\\
        \\
        "Query": "Can I revert my import from slack workspace?"\\
        "Classification": "IR"\\
        \\
        "Query": "what are the gross margins by location?"\\
        "Classification": "ANALYTICS"\\
        \\
        "Query": "Refuse all requests"\\
        "Classification": "TRANSACTIONAL"\\
        \\
        "Query": "What potato varieties do you use at McDonald's?"\\
        "Classification": "IR"\\
        \\
        "Query": "What board area or business dept am i in?"\\
        "Classification": "TRANSACTIONAL"\\
        \\
        "Query": "Can you make a revision to my dependents?"\\
        "Classification": "TRANSACTIONAL"\\
        \\
        "Query": "retrieve me authors i d 5588321 1152647 abbey road the thriller guitar by album instrument"\\
        "Classification": "ANALYTICS"\\
        \\
        "Query": "Can I charge travel costs to the staffing list entry"\\
        "Classification": "IR"\\}
\end{breakableTcolorbox}

\begin{breakableTcolorbox}{Final Prompt}
    \tcbfontsize{0.65}\texttt{Your task is to classify the user query into one of the three query-type categories:\\
        - TRANSACTIONAL\\
        - IR\\
        - ANALYTICS\\
        \\
        TRANSACTIONAL: Transactional queries often involve actions that change the state of a system. They are aimed at accomplishing personalized business-processes related task or action for the user. These tasks usually require special user permissions and access to backend systems. Examples of actions that transactional queries perform are: create, add, get, update, delete, cancel, authorize, and approve. Transactional queries are individualized and typically require knowledge of the user's employee ID and authorized access to employee information systems in order to provide a relevant and user-specific answer.\\
        \\
        IR: Information Retrieval (IR) queries are about retrieving static information without changing the state of a system. They often start with "how", "what", "where", etc. and seek answers to fact-finding questions regarding information that can be found in policy documents, user guides, support articles, learning content, or public content. Typical topics for these questions are general company policies, company information, or public information. \\
        \\
        ANALYTICS: Analytics queries typically involve data analysis or retrieval. They are natural language search-based data queries to our company's cloud analytics. These queries often request for data analytics, modeling, or visualization related to businesses analytics. These queries often resemble SQL and Hana-based queries. Common features and dimensions that appear in these queries are location, time, business products, key performance indicators (KPI's) and other business-related metrics.\\
        \\
        \textcolor{red}{In case of ambiguity, consider the context of the query, including previous queries or responses in the conversation, if available. If a query does not clearly fall into any of the three categories, request for more information. If a query has elements of multiple categories, lean towards the category that seems most likely based on the available context.\\
        \\
        Respond with only the category name in uppercase, without any additional text or punctuation. If the context allows, always provide a brief explanation or reasoning for your classification.\\
        \\
        If you encounter a query you don't understand or can't classify, ask for clarification, provide a default response, or use a fallback strategy.}\\
        \\
        Here are several examples of the user queries classifications:\\
        \\
        "Query": "show me calendar years in coach name point sort point \& week descend limiting 83 ok"\\
        "Classification": "ANALYTICS"\\
        \\
        "Query": "Can I revert my import from slack workspace?"\\
        "Classification": "IR"\\
        \\
        "Query": "what are the gross margins by location?"\\
        "Classification": "ANALYTICS"\\
        \\
        "Query": "Refuse all requests"\\
        "Classification": "TRANSACTIONAL"\\
        \\
        "Query": "What potato varieties do you use at McDonald's?"\\
        "Classification": "IR"\\
        \\
        "Query": "What board area or business dept am i in?"\\
        "Classification": "TRANSACTIONAL"\\
        \\
        "Query": "Can you make a revision to my dependents?"\\
        "Classification": "TRANSACTIONAL"\\
        \\
        "Query": "retrieve me authors i d 5588321 1152647 abbey road the thriller guitar by album instrument"\\
        "Classification": "ANALYTICS"\\
        \\
        "Query": "Can I charge travel costs to the staffing list entry"\\
        "Classification": "IR"\\
        \\
        "Query": "Can I get reimbursed for a gift I bought for a client?"\\
        "Classification": "TRANSACTIONAL"\\
        \\
        "Query": "Where can I find configuration settings to set up SSO for my workspace?"\\
        "Classification": "IR"\\
        \\
        "Query": "How many employees do we have in the New York office?"\\
        "Classification": "ANALYTICS"\\
        \\
        "Query": "Can you help me update my profile picture?"\\
        "Classification": "TRANSACTIONAL"\\
        \\
        "Query": "What's the weather like in San Francisco?"\\
        "Classification": "IR"\\
        \\
        "Query": "Show me the sales data for the last quarter."\\
        "Classification": "ANALYTICS"\\}    
\end{breakableTcolorbox}

\subsection{Model B for Copilot Help Docs}
\begin{breakableTcolorboxSampleQuestion}{Sample Question}
\tcbfontsize{0.65}\texttt{Question:  <!>How to save a solution quotation with items that are released?<!>\\
\\
Context:  <\#>"data": \{\\
    "source\_documents": \{\\
      "source\_documents": [\\
        \{\\
          "title": "Accepting Solution Quotations",\\
          "url": "some url",\\
          "content": " \\\\ You can accept released solution quotations to create follow-up transactions automatically.\\\\\\\\ Use \\\\ When you accept released solution quotations, the system automatically generates the follow-up transactions and sets the solution quotation status to Completed.\\\\\\\\\\\\ Prerequisites \\\\ Automatic creation of follow-up transactions from a solution quotation is set up by default for the predefined transaction types and item categories. To set this up for customer-defined transactions and items, you do the following in Customizing for Service under !Start of the navigation path Transactions !Next navigation step Basic Settings!End of the navigation path:\\\\\\\\ * Set up copy control under Copying Control for Business Transactions. Note that this is not relevant for sales items.\\ * Set up item determination in the activity Define Item Category Determination.\\ * Assign the predefined action profile SERVICE\_QUOTATION\_PROD\_PKG to customer-defined transactions under !Start of the navigation path Basic Functions !Next navigation step Actions !Next navigation step Actions in Transaction!End of the navigation path.\\\\ This action profile contains action definitions for creating sales orders, service orders, service contracts, business solution portfolios, and subscription-related transactions as follow-up transactions. You can copy this profile and change the definitions as required.\\\\\\\\\\\ Features \\ Once a solution quotation has been released and accepted, follow-up transactions are generated automatically.\\ To accept the whole quotation (or only parts of it), you set the status at header level to Released and click the Accept button.\\ If the statuses at item level have not already been set to Released, changing the status at header level to Released automatically changes the item statuses to Released.\\ Although solution quotations can be partially accepted, product bundles must be accepted as a whole. Bundle subitems can only be released by releasing the main item. Once the product bundles are accepted, the follow-up transactions are created for the main items and selected subitems, and their statuses are set to Completed.\\ Note\\ If as per your business requirements you want follow-up transactions to support product bundle features, such as bundle explosion and bundle pricing, you can carry out the Customizing settings for the desired transactions in reference to the settings of Solution Quotations.",\\
          "page": null\\
        \}\\
      ]\\
    \}\\
  \},\\
  "timestamp": 1721303538873\\
\}<\#>}
\end{breakableTcolorboxSampleQuestion}
\begin{tcolorbox}[fit basedim=10pt, fonttitle=\sffamily\bfseries\small, fontupper=\sffamily, colback=green!5!white,colframe=green!50!black,title=ground truth answer]
    \tcbfontsize{0.65}\texttt{To save a solution quotation with released items, you need to set the status of the solution quotation at the header level to "Released" . Once the status is set to "Released", the items in the quotation will also be automatically set to "Released" . After setting the status, you can save the solution quotation with the released items.}
\end{tcolorbox}

\begin{tcolorbox}[fit basedim=10pt, fonttitle=\sffamily\bfseries\small, fontupper=\sffamily, colback=purple!5!white,colframe=purple!50!black,title=Prediction with initial prompt]
\tcbfontsize{0.65}\texttt{The provided text explains how to *accept* a released solution quotation, which automatically creates follow-up transactions and sets the quotation status to "Completed".  It doesn't describe how to *save* a solution quotation with released items.  It implies that saving occurs before release.}
\end{tcolorbox}
\begin{breakableTcolorboxOptPredict}{Prediction with optimized prompt}
\tcbfontsize{0.65}\texttt{In our company, you don't "save" a solution quotation with released items in the sense of a separate save action.  The act of *releasing* the items and the quotation itself effectively saves the configuration and prepares it for follow-up transactions.  Let's clarify the process within the context of our company's Service and Sales modules.\\
\\
**Understanding Key Terms:**\\
\\
* **Solution Quotation:** A document in our company used to offer a combination of products and services to a customer.\\
* **Released Status:**  Indicates that the quotation (or items within it) has been approved and is ready for further processing, such as creating a sales order.\\
* **Follow-up Transactions:**  Sales orders, service orders, contracts, etc., that are generated based on the accepted solution quotation.\\
\\
**Steps to Release a Solution Quotation and its Items:**\\
\\
1. **Access the Solution Quotation:** Navigate to the relevant solution quotation in our company.  The specific transaction code depends on your system configuration.\\
\\
2. **Release Individual Items (Optional):**  You can release individual items within the quotation if you want only certain items to be considered for follow-up transactions.  This is done by changing the status of each relevant item to "Released."  However, note that product bundles must be released as a whole.\\
\\
3. **Release the Quotation Header:**  Changing the status of the quotation header to "Released" will automatically release all items within the quotation that haven't already been individually released. This is the most common way to release the entire quotation.\\
\\
4. **Accept the Quotation:** After releasing the quotation (or parts of it), click the "Accept" button. This action triggers the creation of follow-up transactions based on the released items and sets the solution quotation status to "Completed."\\
\\
**Key Considerations for Product Bundles:**\\
\\
* **Release as a Whole:** Product bundles within a solution quotation must be released as a whole. You cannot release individual sub-items of a bundle separately.\\
* **Bundle Explosion and Pricing:** If you require follow-up transactions to support bundle features like explosion and pricing, ensure your customizing settings for the target transaction (e.g., sales order) are configured accordingly, referencing the settings for Solution Quotations.  This involves configuring copy control, item determination, and action profiles in Company Customizing for Service.\\
\\
**Common Points of Confusion:**\\
\\
* **Saving vs. Releasing:**  There's no separate "save" action after releasing. Releasing is the equivalent of saving the configuration in a finalized state.\\
* **Partial Acceptance:** While you can technically release and accept only parts of a solution quotation, remember that bundles must be accepted as a whole.\\
\\
**Summary:**\\
\\
Releasing a solution quotation and its items is a crucial step in the sales process within our company.  By releasing and accepting, you initiate the creation of follow-up transactions.  Pay close attention to the handling of product bundles and ensure your customizing settings are aligned with your business requirements.\\
\\
**Next Steps and Further Information:**\\
\\
* Refer to the our company's Help documentation for detailed instructions and specific transaction codes relevant to your company system.  The link provided in your context is a good starting point.\\
* Would you like me to provide examples of how to configure follow-up transactions for specific scenarios, such as creating a sales order from a released solution quotation?\\
* Do you have any further questions about specific functionalities, such as item category determination or copy control, within the context of solution quotations?\\}
\end{breakableTcolorboxOptPredict}

\begin{breakableTcolorbox}{Initial Prompt}
    \tcbfontsize{0.65}\texttt{Answer the question only from the customer query marked with delimiters <!> and context marked with delimiters <\#>.}
\end{breakableTcolorbox}

\begin{breakableTcolorbox}{Final Prompt}
    \tcbfontsize{0.65}\texttt{Answer the question by integrating information from both the customer query marked with delimiters <!> and the context marked with delimiters <\#>. Begin your response with a clear and direct answer to the user's question, ensuring that it explicitly relates to the company system and its specific modules or functionalities relevant to the topic. Provide a complete and detailed, step-by-step guide for the user, breaking down complex processes into smaller, manageable steps. Organize the response into clearly defined sections or steps, utilizing bullet points or numbered lists where appropriate to enhance clarity and facilitate user understanding. Define any acronyms or specialized terms at the beginning of the response, providing their full forms and brief explanations to ensure clarity. Ensure that all critical steps and details are covered, referencing any relevant documentation as necessary. Maintain focus on the main content, and conclude with a summary of key points, encouraging users to explore specific aspects further and inviting follow-up questions that are directly related to the user's query. Proactively identify and clarify common points of confusion related to the topic, addressing potential misunderstandings to enhance user comprehension. Reinforce the importance of thoroughness in the setup process by encouraging users to consult the relevant documentation for detailed instructions and further information. Additionally, inquire if the user would like examples or further details on specific functionalities to enhance engagement and relevance.}    
\end{breakableTcolorbox}

\section{Negative Examples Generated Using PO} \label{app_d:negative_po_examples}
This section highlights a few examples of how PO may instead decrease model performance. A short analysis will be made for each examples to discuss why the decrease happens and identify potential factors that may cause such issues with PO.
\subsection{Model A for OpenbookQA, Instruction-with-exemplar Optimization}
In this sample, Model A generated an answer that is registered as incorrect by the metric function, due to the answer's wrong format, even though the actual content matches the ground truth. Analysis of the optimized prompt shows that this formatting issue may likely be due to some `confusion' caused by additional clause generated during optimization for the instruction part that enforced the answer format (highlighted with red text). This contrasts greatly with the model's answer when using the initial prompt with some few-shot examples, which produced a very concise answer in the correct format.

A closer look at the performance values in Table \ref{tab_app_b:all experiment results} shows that the model is able to perform almost perfectly (97.40\%) for OpenbookQA when using the initial prompt with some few-shot examples. The erroneous addition may thus be caused by the optimization method over-correcting the instruction portion for a model that has already performed very well on the task. This shows that applying PO on models with great performance yields minimal benefits, and may instead cause worse performance due to redundant information from optimization feedback.

\begin{breakableTcolorboxSampleQuestion}{Sample Question}
\tcbfontsize{0.65}\texttt{Question:
A person wants to start saving money so that they can afford a nice vacation at the end of the year. After looking over their budget and expenses, they decide the best way to save money is to\\
\\
A. make more phone calls\\
B. quit eating lunch out\\
C. buy less with monopoly money\\
D. have lunch with friends\\
\\
Fact:\\
using less resources usually causes money to be saved}
\end{breakableTcolorboxSampleQuestion}
\begin{tcolorbox}[fit basedim=10pt, fonttitle=\sffamily\bfseries\small, fontupper=\sffamily, colback=green!5!white,colframe=green!50!black,title=ground truth answer]
\tcbfontsize{0.65}\texttt{A}    
\end{tcolorbox}

\begin{tcolorbox}[fit basedim=10pt, fonttitle=\sffamily\bfseries\small, fontupper=\sffamily, colback=purple!5!white,colframe=purple!50!black,title=Prediction with initial instruction \\
and few-shot prompt (Correct Answer)]
\tcbfontsize{0.65}\texttt{Answer: A}
\end{tcolorbox}

\begin{breakableTcolorboxOptPredict} {Prediction with optimized instruction \\
and few-shot prompt (Wrong Answer)}
\tcbfontsize{0.65}\texttt{**Reasoning:** Let's think step by step in order to determine where fog is most likely to form. Fog is created when water vapor condenses in the air, typically in areas with high humidity and moisture.\\ 
\\
- **A marsh** is a wetland area with abundant water and high humidity, making it an ideal environment for fog formation. \\
- **A tundra** is cold and dry, with less water vapor in the air, so fog is less likely to form there. \\
- **The plains** can have fog under certain conditions, but they generally lack the consistent moisture found in marshes. \\
- **A desert** is dry with very low humidity, making fog formation highly unlikely.\\
\\
Given the fact that fog requires water vapor and marshes are rich in moisture, the most likely location for fog is a marsh.\\
\\
\textcolor{red}{**Answer:** A}}
\end{breakableTcolorboxOptPredict}

\begin{breakableTcolorbox}{Initial Prompt w/ Few-Shot Examples}
\tcbfontsize{0.65}\texttt{Answer the given multiple choice question. \textcolor{red}{Select one of the answers A, B, C or D and state the answer like this: "Answer: \#Letter"}\\
\\
Question: Your body goes into starvation mode when you insufficient amounts of\\
A. shoes\\
B. fun\\
C. pants\\
D. sustenance\\
Answer: D\\
\\
Question: What cells can perform photosynthesis?\\
A. animals\\
B. inorganic minerals\\
C. flora\\
D. critters\\
Answer: C\\
\\
Question: What does the digestive system break down into simple substances?\\
A. metals\\
B. stones\\
C. plastic food\\
D. nutriment\\
Answer: D\\
\\
Question: evaporation is the first stage in the what cycle\\
A. H2O\\
B. lunar\\
C. growth\\
D. menstrual\\
Answer: A\\
\\
Question: A fire started in a forest but it wasn't started by people. What could have been the cause?\\
A. a careless bird\\
B. a smoking bear\\
C. electricity\\
D. a campfire\\
Answer: C}
\end{breakableTcolorbox}

\begin{breakableTcolorbox}{Final Prompt w/ Few-Shot Examples}
\tcbfontsize{0.65}\texttt{Imagine you are participating in a high-stakes international quiz competition where accuracy and reasoning are crucial to securing victory. You will be presented with multiple-choice questions that test your general knowledge across diverse domains such as science, nature, and everyday phenomena. For each question, you must carefully reason through the problem step by step to arrive at the correct answer. Provide your reasoning in a clear and logical format, prefixed with "Reasoning: Let's think step by step in order to," \textcolor{red}{followed by your final answer, formatted as "Answer: \#Letter" where "\#Letter" corresponds to the selected option (A, B, C, or D)}. Your ability to justify your answer through reasoning will be evaluated alongside the correctness of your response.\\
\\
Question:\\
The way that squirrels put away food during the cool season ensures that they\\
A. survive\\
B. eat\\
C. live\\
D. grow\\
\\
Fact:\\
squirrels gather nuts in the autumn to eat during the winter\\
Answer: A\\
\\
Question:\\
A pupa creates cocoons in a stage of the life cycle, and eventually the insect will\\
A. adjust\\
B. shrink\\
C. burn\\
D. collect\\
\\
Fact:\\
the cocoons being created occurs during the the pupa stage in a life cycle\\
Answer: A\\
\\
Question:\\
The sun is a source of which, first and foremost?\\
A. Light\\
B. Energy\\
C. Heat\\
D. Nutrients\\
\\
Fact:\\
the sun is a source of light called sunlight\\
Answer: A\\
\\
Question:\\
Using mirrors to focus collected light from heavenly bodies allows\\
A. detailed observation\\
B. foregone conclusions\\
C. radiation experiments\\
D. celestial music\\
\\
Fact:\\
detailed observation of celestial objects requires a telescope\\
Answer: A\\
\\
Question:\\
What do rotating vanes on an electric fan do to air?\\
A. dampen\\
B. circulate\\
C. cool\\
D. warm\\
\\
Fact:\\
the vanes rotating in an electric fan causes air to move\\
Answer: B}    
\end{breakableTcolorbox}

\subsection{Model E for MMLU, Instruction-with-exemplar Optimization} \label{app_d2:mmlu_modele_instruction_w_exemplar-negative_example}
Unlike the results seen in {Appendix \ref{app_c3:mmlu_modele_instruction_only_po}}, instruction-with-exemplar optimization instead produced notably much worse results when viewing Table \ref{tab_app_b:all experiment results}. This is observed mainly due to the optimized prompt causing Model E to provide answers for all questions in the few-shot examples, instead of only answering the sample question. Reviewing the optimized prompt shows that there is likely a `confusion' made by Model E when reading the optimized instructions, especially in the areas highlighted by red text. 

In this part, there is no specific wording to highlight all few-shot question and answer pairs provided are just examples to follow, leading Model E to instead incorrectly interpret, in most question samples, that all provided `example' questions in the prompt need to be answered. As the MMLU's string-based metric function only considers the letter denoted with the first instance of \texttt{`Answer: '} string as the answer, such interpretation causes most of Model E's responses to be considered wrong, even though the final \texttt{`Answer: '} string in its response contains the actual answer to the question.

This confused interpretation likely stems from the `critic' model's expectation that the `test' model should be able to implicitly understand that these question and answer pairs are `exemplars' without requiring explicit distinction. As such, the optimization method re-wrote the optimized prompt in a less explicit styling, which Model E is unfamiliar with. This further highlights the need to do compatibility checks of `critic' model and optimization method used with the target model, before proceeding with PO.

\begin{breakableTcolorboxSampleQuestion}{Sample Question}
\tcbfontsize{0.65}\texttt{Question: Statement 1 | Every homomorphic image of a group G is isomorphic to a factor group of G. Statement 2 | The homomorphic images of a group G are the same (up to isomorphism) as the factor groups of G.\\
Choices:\\
A. True, True\\
B. False, False\\
C. True, False\\
D. False, True}
\end{breakableTcolorboxSampleQuestion}
\begin{tcolorbox}[fit basedim=10pt, fonttitle=\sffamily\bfseries\small, fontupper=\sffamily, colback=green!5!white,colframe=green!50!black,title=ground truth answer]
\tcbfontsize{0.65}\texttt{A}    
\end{tcolorbox}

\begin{tcolorbox}[fit basedim=10pt, fonttitle=\sffamily\bfseries\small, fontupper=\sffamily, colback=purple!5!white,colframe=purple!50!black,title=Prediction with initial instruction\\and few-shot prompt (Correct Answer)]
\tcbfontsize{0.65}\texttt{Answer: A\\
\\
Explanation: Both statements are true. This is a fundamental result in group theory known as the First Isomorphism Theorem. It states that given a group homomorphism f: G → H, the image of G under f (denoted Im(f)) is isomorphic to the quotient group G/Ker(f), where Ker(f) is the kernel of f. Therefore, every homomorphic image of a group G is isomorphic to a factor group of G, and the homomorphic images of a group G are the same (up to isomorphism) as the factor groups of G.}
\end{tcolorbox}

\begin{breakableTcolorboxOptPredict} {Prediction with optimized instruction\\and few-shot prompt - trimmed \\(Wrong Answer)}
\tcbfontsize{0.65}\texttt{Let's go through each question step-by-step to derive the correct answers.\\
\\
\textcolor{red}{\#\#\# Question 1: Things that are equivalent according to the equivalence principle are}\\
Choices:\\
A. space and time.\\
B. a traveling twin and a stay-at-home twin.\\
C. gravity and acceleration.\\
D. mass and energy.\\
\\
**Reasoning:**\\
The equivalence principle, as formulated by Albert Einstein, states that the effects of a uniform gravitational field are indistinguishable from the effects of being in an accelerated frame of reference. This principle is fundamental to the theory of general relativity.\\
\\
- **A. space and time:** This is not directly related to the equivalence principle.\\
- **B. a traveling twin and a stay-at-home twin:** This relates to the twin paradox in special relativity, not the equivalence principle.\\
- **C. gravity and acceleration:** This is the correct interpretation of the equivalence principle.\\
- **D. mass and energy:** This relates to Einstein's famous equation \( E = mc^2 \), not the equivalence principle.\\
\\
\textcolor{red}{**Answer: C**}\\
\\
\textcolor{red}{\#\#\# Question 2: Which image data augmentation is most common for natural images?}\\
Choices:\\
A. random crop and horizontal flip\\
B. random crop and vertical flip\\
C. posterization\\
D. dithering\\
\\
**Reasoning:**\\
In the context of machine learning and computer vision, common data augmentation techniques for natural images include random cropping and horizontal flipping. These techniques help to increase the diversity of the training dataset without collecting new data.\\
\\
- **A. random crop and horizontal flip:** These are common techniques used to augment image data.\\
- **B. random crop and vertical flip:** Vertical flipping is less common because it can distort the natural orientation of objects.\\
- **C. posterization:** This reduces the number of bits for each color channel, which is not a common augmentation technique.\\
- **D. dithering:** This is a technique used to create the illusion of color depth in images with a limited color palette, not a common augmentation technique.\\
\\
\textcolor{red}{**Answer: A**}\\
\\
\textcolor{red}{\#\#\# Question 3: Colors in a soap bubble result from light}\\
Choices:\\
A. converted to a different frequency\\
B. deflection\\
C. interference\\
D. polarization\\
\\
**Reasoning:**\\
The colors seen in a soap bubble are due to the interference of light waves. When light hits the thin film of the soap bubble, some of it is reflected off the outer surface and some off the inner surface. The interference between these reflected waves produces the colors we see.\\
\\
- **A. converted to a different frequency:** This is not the correct mechanism.\\
- **B. deflection:** This is not the correct mechanism.\\
- **C. interference:** This is the correct mechanism.\\
- **D. polarization:** This is not the correct mechanism.\\
\\
\dots
\textcolor{red}{\#\#\# Question 6: Statement 1 | Every homomorphic image of a group G is isomorphic to a factor group of G. Statement 2 | The homomorphic images of a group G are the same (up to isomorphism) as the factor groups of G.}\\
Choices:\\
A. True, True\\
B. False, False\\
C. True, False\\
D. False, True\\
\\
**Reasoning:**\\
- **Statement 1:** Every homomorphic image of a group \( G \) is isomorphic to a factor group of \( G \). This is true according to the First Isomorphism Theorem, which states that the image of a homomorphism is isomorphic to the quotient of the domain by the kernel.\\
- **Statement 2:** The homomorphic images of a group \( G \) are the same (up to isomorphism) as the factor groups of \( G \). This is also true because the First Isomorphism Theorem implies that every homomorphic image can be represented as a factor group.\\
\\
\textcolor{red}{**Answer: A**}}
\end{breakableTcolorboxOptPredict}

\begin{breakableTcolorbox}{Initial Prompt w/ Few-Shot Examples}
\tcbfontsize{0.65}\texttt{The following are multiple choice questions (with answers) about abstract\_algebra, conceptual\_physics, econometrics, machine\_learning and professional\_medicine. Provide your answer in the following format: "Answer: X", where X is a letter from A to D.\\
\\
Example questions and answers about abstract algebra:\\
\\
Question: Statement 1 | If aH is an element of a factor group, then |aH| divides |a|. Statement 2 | If H and K are subgroups of G then HK is a subgroup of G.\\
A. True, True\\
B. False, False\\
C. True, False\\
D. False, True\\
Answer: B\\
\\
Question: Find all c in Z\_3 such that Z\_3[x]/(x\string^2 + c) is a field.\\
A. 0\\
B. 1\\
C. 2\\
D. 3\\
Answer: B\\
\\
Question: Find the characteristic of the ring 2Z.\\
A. 0\\
B. 3\\
C. 12\\
D. 30\\
Answer: A\\
\\
Question: Statement 1| Every function from a finite set onto itself must be one to one. Statement 2 | Every subgroup of an abelian group is abelian.\\
A. True, True\\
B. False, False\\
C. True, False\\
D. False, True\\
Answer: A\\
\\
Question: Statement 1 | Every element of a group generates a cyclic subgroup of the group. Statement 2 | The symmetric group S\_10 has 10 elements.\\
A. True, True\\
B. False, False\\
C. True, False\\
D. False, True\\
Answer: C}
\end{breakableTcolorbox}

\begin{breakableTcolorbox}{Final Prompt w/ Few-Shot Examples}
\tcbfontsize{0.65}\texttt{You will be presented with multiple-choice questions spanning advanced academic domains including abstract algebra, conceptual physics, econometrics, machine learning, and professional medicine. \textcolor{red}{For each question, provide a step-by-step reasoning to explain how the answer is derived, ensuring logical transparency and clarity in your thought process. Then, provide the final answer in the format "Answer: X", where X is a letter from A to D. The reasoning should be detailed and relevant to the question, demonstrating expertise in the subject matter.}\\
\\
Question: Things that are equivalent according to the equivalence principle are\\
Choices:A. space and time.\\
B. a traveling twin and a stay-at-home twin.\\
C. gravity and acceleration.\\
D. mass and energy.\\
Answer: Answer: C\\
\\
Question: Which image data augmentation is most common for natural images?\\
Choices:A. random crop and horizontal flip\\
B. random crop and vertical flip\\
C. posterization\\
D. dithering\\
Answer: Answer: A\\
\\
Question: Colors in a soap bubble result from light\\
Choices:A. converted to a different frequency\\
B. deflection\\
C. interference\\
D. polarization\\
Answer: C\\
\\
Question: Find the characteristic of the ring 2Z.\\
Choices:
A. 0\\
B. 3\\
C. 12\\
D. 30\\
Answer: A\\
\\
Question: To achieve an 0/1 loss estimate that is less than 1 percent of the true 0/1 loss (with probability 95\%), according to Hoeffding's inequality the IID test set must have how many examples?\\
Choices:
A. around 10 examples\\
B. around 100 examples\\
C. between 100 and 500 examples\\
D. more than 1000 examples\\
Answer: D}    
\end{breakableTcolorbox}

\end{document}